\documentclass{article}

\usepackage{microtype}
\usepackage{graphicx}
\usepackage{subfigure}
\usepackage{booktabs} 
\usepackage[table]{xcolor}

\usepackage{hyperref}

\usepackage[accepted]{icml2023}

\usepackage{amsmath}
\usepackage{amssymb}
\usepackage{mathtools}
\usepackage{amsthm}

\theoremstyle{plain}

\theoremstyle{definition}

\theoremstyle{remark}

\usepackage[textsize=tiny]{todonotes}

\usepackage{multirow}
\usepackage{microtype}
\usepackage{booktabs}
\usepackage{enumitem}
\usepackage[utf8]{inputenc}
\usepackage{listings}
\usepackage{caption}
\usepackage{dsfont}
\usepackage{graphicx}
\usepackage{algorithm}
\usepackage{amsthm}
\usepackage{afterpage}
\usepackage{xspace}
\usepackage{adjustbox}

\usepackage{ragged2e} 
\usepackage{booktabs, tabularx}
\usepackage{makecell}
\usepackage{nicematrix,tikz}
\usepackage{wrapfig}
\usepackage{appendix}

\usepackage{colortbl}
\usepackage{pifont}
\usepackage{xspace}
\usepackage{soul}
\usepackage{etoc}

\makeatletter
\def\part{\par
   \addvspace{4ex}%
   \@afterindentfalse
   \secdef\@part\@spart}%
\def\@part[#1]#2{%
 \@ifnum{\c@secnumdepth >\m@ne}{%
        \refstepcounter{part}%
        \addcontentsline{toc}{part}{\thepart\hspace{1em}#1}%
 }{%
      \addcontentsline{toc}{part}{#1}%
 }%
   \nobreak
   \vskip 3ex
   \@afterheading
}%
\makeatother

\newcommand{\ms}[2]{{#1\tiny{$\pm$#2}}}

\newcommand*{\scale}[2][4]{\scalebox{#1}{$#2$}}

\usepackage[capitalize,noabbrev,nameinlink]{cleveref}
\definecolor{mydarkblue}{rgb}{0,0.08,0.45}
\definecolor{myblue}{HTML}{3b75c3}
\definecolor{myred}{HTML}{E33222}
\definecolor{mygreen}{HTML}{438773}
\definecolor{mymaroon}{RGB}{142,27,19}
\definecolor{maroon}{HTML}{800000}
\definecolor{mycite}{cmyk}{0.55,1,0,0.15}
\definecolor{codeblue}{rgb}{0.25,0.5,0.5}
\definecolor{codekw}{rgb}{0.85, 0.18, 0.50}
\definecolor{codegreen}{rgb}{0,0.6,0}
\definecolor{codegray}{rgb}{0.5,0.5,0.5}
\definecolor{codepurple}{rgb}{0.58,0,0.82}
\definecolor{backcolour}{rgb}{0.95,0.95,0.92}
\hypersetup{
    colorlinks=true,
    citecolor=mycite,
    urlcolor=mydarkblue,
    linkcolor=codekw
}

\newcommand{\kdp}{$\mathcal{L}_{LM}$\xspace}
\newcommand{\kdf}{$\mathcal{L}_{RM}$\xspace}
\newcommand{\lsup}{$\mathcal{L}_{sup}$\xspace}

\newcommand{\ours}{\textsf{LLP}\xspace}
\newcommand{\kdr}{$\mathcal{L}_{\ours\_R}$\xspace}
\newcommand{\kdkl}{$\mathcal{L}_{\ours\_D}$\xspace}

\newcommand{\cora}{\texttt{Cora}\xspace}
\newcommand{\citeseer}{\texttt{Citeseer}\xspace}
\newcommand{\pubmed}{\texttt{Pubmed}\xspace}
\newcommand{\cs}{\texttt{CS}\xspace}
\newcommand{\physics}{\texttt{Physics}\xspace}
\newcommand{\computers}{\texttt{Computers}\xspace}
\newcommand{\photos}{\texttt{Photos}\xspace}
\newcommand{\collab}{\texttt{Collab}\xspace}
\newcommand{\citationtwo}{\texttt{Citation2}\xspace}
\newcommand{\igbtiny}{\texttt{IGB-100K}\xspace}
\newcommand{\igbsmall}{\texttt{IGB-1M}\xspace}

\usepackage{amsmath,amsfonts,bm}

\def\eqref#1{equation~\ref{#1}}

\def\1{\bm{1}}

\def\mA{{\bm{A}}}

\DeclareMathAlphabet{\mathsfit}{\encodingdefault}{\sfdefault}{m}{sl}
\SetMathAlphabet{\mathsfit}{bold}{\encodingdefault}{\sfdefault}{bx}{n}

\icmltitlerunning{Linkless Link Prediction via Relational Distillation}

\begin{document}

\twocolumn[
\icmltitle{Linkless Link Prediction via Relational Distillation}



\icmlsetsymbol{equal}{*}

\begin{icmlauthorlist}
\icmlauthor{Zhichun Guo}{equal,nd}
\icmlauthor{William Shiao}{ucr}
\icmlauthor{Shichang Zhang}{ucla}
\icmlauthor{Yozen Liu}{snap}
\icmlauthor{Nitesh V. Chawla}{nd}
\icmlauthor{Neil Shah}{snap}
\icmlauthor{Tong Zhao}{snap}
\end{icmlauthorlist}

\icmlaffiliation{nd}{Department of Computer Science and Engineering, University of Notre Dame, IN, USA}
\icmlaffiliation{ucr}{Department of Computer Science and Engineering, University of California, Riverside, CA, USA}
\icmlaffiliation{ucla}{Department of Computer Science, University of California, Los Angeles, CA, USA}
\icmlaffiliation{snap}{Snap Inc., CA, USA}

\icmlcorrespondingauthor{Zhichun Guo}{zguo5@nd.edu}


\vskip 0.3in
]



\printAffiliationsAndNotice{\icmlEqualContribution} 

\begin{abstract}
Graph Neural Networks (GNNs) have shown exceptional performance in the task of link prediction. Despite their effectiveness, the high latency brought by non-trivial neighborhood data dependency limits GNNs in practical deployments. Conversely, the known efficient MLPs are much less effective than GNNs due to the lack of relational knowledge. In this work, to combine the advantages of GNNs and MLPs, we start with exploring direct knowledge distillation (KD) methods for link prediction, i.e., predicted logit-based matching and node representation-based matching. Upon observing direct KD analogs do not perform well for link prediction, we propose a relational KD framework, \emph{Linkless Link Prediction} (\ours), to distill knowledge for link prediction with MLPs. Unlike simple KD methods that match independent link logits or node representations, \ours distills \emph{relational} knowledge that is centered around each (anchor) node to the student MLP. Specifically, we propose rank-based matching and distribution-based matching strategies that complement each other.
Extensive experiments demonstrate that \ours boosts the link prediction performance of MLPs with significant margins, and even outperforms the teacher GNNs on 7 out of 8 benchmarks. \ours also achieves a 70.68$\times$ speedup in link prediction inference compared to GNNs on the large-scale OGB dataset.
\end{abstract}

\section{Introduction}
Graph neural networks (GNNs) have been widely used for machine learning on graph-structured data~\citep{kipf2016semi, hamilton2017inductive}. They have shown significant performance in various applications, such as node classification~\citep{velivckovic2017graph, chen2020simple}, graph classification~\citep{zhang2018end}, graph generation~\citep{you2018graphrnn,shiao2021adversarially}, and link prediction~\citep{zhang2018link,shiao2022link}.

Of these, link prediction is a notably critical problem in the graph machine learning community, which aims to predict 
the likelihood of any two nodes forming a link.
It has broad practical applications such as knowledge graph completion~\citep{schlichtkrull2018modeling,nathani2019learning,vashishth2020composition}, friend recommendation on social platforms~\citep{sankar2021graph, tang2022friend,fan2022graph} and item recommendation for users on service and commerce platforms~\citep{koren2009matrix,ying2018graph, he2020lightgcn}. 
With the rising popularity of GNNs, state-of-the-art link prediction methods adopt encoder-decoder style models, where encoders are GNNs, and decoders are applied directly on pairs of node representations learned by the GNNs~\citep{kipf2016variational,zhang2018link, cai2020multi,zhao2022learning}. 

The success of GNNs is typically attributed to the explicit use of contextual information from nodes' surrounding neighborhoods~\citep{zhang2020deep}. However, this induces a heavy reliance on neighborhood fetching and aggregation schemes, which can lead to high time cost in training and inference compared to tabular models, 
such as multi-layer perceptrons (MLPs), especially owing to neighbor explosion~\citep{zhang2020agl, jia2020redundancy,  zhang2021graph, zeng2019graphsaint}. 
Compared to GNNs, MLPs do not require any graph topology information, making them more suitable for new or isolated nodes (e.g., for cold-start settings), but usually resulting in worse general task performance as encoders, which we also empirically validate in \cref{sec:exp}. 
Nonetheless, having no graph dependency makes the training and inference time for MLPs negligible when comparing with those of GNNs.
Thus, in large-scale applications where fast real-time inference is required, MLPs are still a leading option \citep{zhang2021graph, covington2016deep, gholami2021survey}. 

Given these speed-performance tradeoffs, several recent works propose to transfer the learned knowledge from GNNs to MLP using knowledge distillation (KD) techniques~\citep{hinton2015distilling, zhang2021graph, zheng2021cold, hu2021graph}, to take advantage of both GNN's performance 
benefits and MLP's speed benefits. Specifically, in this way, 
the student MLP can potentially obtain the graph-context knowledge transferred from the GNN teacher via KD to not only perform better in practice,
but also enjoy model latency benefits compared to GNNs, e.g. in production inference settings. However, these works focus on node- or graph-level tasks. Given that KD on link prediction tasks have not been explored,
and the massive scope of recommendation systems contexts that are posed as link prediction problems, our work aims to bridge a critical gap.  Specifically, we ask:

\begin{center}
\vspace{-5pt}
\textbf{\textit{Can we effectively distill link prediction-relevant knowledge from GNNs to MLPs?}}
\vspace{-5pt}
\end{center}

In this work, we focus on exploring, building upon, and proposing cross-model (GNN to MLP) distillation techniques for link prediction settings. We start with exploring two direct KD methods of aligning student and teacher: (i) logit-based matching of predicted link existence probabilities~\citep{hinton2015distilling}, and (ii) representation-based matching of the generated latent node representations~\citep{gou2021knowledge}. 
However, empirically we observe that neither the logit-based matching nor the representation-based matching are powerful enough to distill sufficient knowledge for the student model to perform well on link prediction tasks.
We hypothesize that the reason of these two KD approaches not performing well is that link prediction, unlike node classification, heavily relies on \emph{relational} graph topological information~\citep{martinez2016survey,zhang2018link,yun2021neo,zhao2022learning}, which is not well-captured by direct methods.

To address this issue, we propose a relational KD framework, namely \ours: our key intuition is that instead of focusing on matching individual node pairs or node representations, we focus on matching the relationships between each (anchor) node with respect to other (context) nodes in the graph.
Given the relational knowledge centered at the anchor node, i.e., the teacher model's predicted link existence probabilities between the anchor node and each context node, \ours distills it to the student model via two matching methods: (i) rank-based matching, and (ii) distribution-based matching. 
More specifically, rank-based matching equips the student model with a ranking loss over the relative ranks of all context nodes w.r.t the anchor node, preserving crucial ranking information that are directly relevant to downstream link prediction use-cases,
e.g. user-contextual friend recommendation~\citep{sankar2021graph, tang2022friend} or item recommendation~\citep{ying2018graph, he2020lightgcn}.
On the other hand, distribution-based matching equips the student model with the link probability distribution over context nodes, conditioned on the anchor node. 
Importantly, distribution-based matching is complementary to rank-based matching, as it provides auxiliary information about the relative values of the probabilities and magnitudes of differences.
To comprehensively evaluate the effectiveness of our proposed \ours, we conduct experiments on 8 public benchmarks. In addition to the standard transductive setting for graph tasks, we also design a more realistic setting that mimics realistic (on-line) use-cases for link prediction, which we call the production setting. 
\ours consistently outperforms stand-alone 
MLPs by 18.18 points on average under the transductive setting and 12.01 points under the production setting on all the datasets, and matches or outperforms teacher GNNs on \textbf{7/8} datasets under the transductive setting. 
Promisingly, for cold-start nodes, \ours outperforms teacher GNNs and stand-alone MLPs by \textbf{25.29} and \textbf{9.42} Hits@20 on average, respectively.
Finally, \ours infers drastically faster than GNNs, e.g. \textbf{70.68$\times$} faster on the large-scale \collab dataset.

\section{Related Work and Preliminaries}
\label{sec: prelimi}
We briefly discuss related work and preliminaries relevant to contextualizing our methods and contributions. Due to space limit, we defer more related work to \cref{sec:related}.

\textbf{Notation.} Let $G = (\mathcal{V}, \mathcal{E})$ denote an undirected graph, where $\mathcal{V}$ denotes the set of $N$ nodes and $\mathcal{E} \subseteq \mathcal{V} \times \mathcal{V}$ denotes the set of observed links. $\textbf{A} \in \{0, 1\} ^ {N \times N}$ denotes the adjacency matrix, where $\textbf{A}_{i,j} = 1$ if exists an edge $e_{i,j}$ in $\mathcal{E}$ and $0$ otherwise. Let the matrix of node features be denoted by $\textbf{X} \in \mathbb{R} ^ {N \times F}$, where each row $\boldsymbol{x}_i$ is the $F$-dim raw feature vector of node $i$. Given both $\mathcal{E}$ and $\mathbf{A}$ have the validation and test links masked off for link prediction, we use $a_{i,j}$ (different from $\mA_{i,j}$) to denote the true label of link existence of nodes $i$ and $j$, which may or may not be visible during training depending on the setting. We use $\mathcal{E}^{-} = (\mathcal{V} \times \mathcal{V}) \setminus \mathcal{E}$ to denote the no-edge node pairs that are used as negative samples during model training. We denote node representations by $\textbf{H} \in \mathbb{R} ^ {N \times D}$, where $D$ is the hidden dimension.  In KD context with multiple models, we use $\boldsymbol{h}_i$ and $\hat{\boldsymbol{h}}_i$ to denote node $i$'s representations learned by the teacher and student models, respectively.  Similarly, we use ${y}_{i,j}$ and $\hat{y}_{i,j}$ to denote the predictions for $a_{i,j}$ by the teacher and the student models, respectively. 

\begin{figure*}[t]
    \centering
    \includegraphics[width=0.97\textwidth]{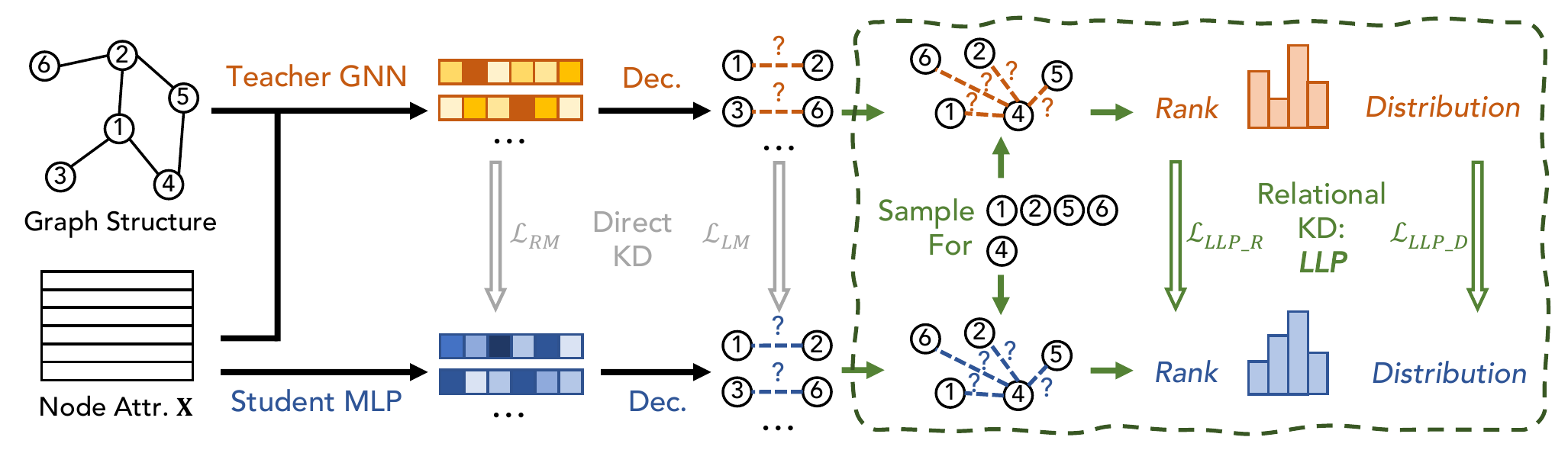}
    \caption{
    We explore KD methods for link prediction, which distill knowledge from a {\color{orange} Teacher GNN} to a {\color{myblue} Student MLP} encoder, each with their own decoder. We start by exploring {\color{codegray} direct KD methods: representation-matching and logit-matching} (\cref{sec:directkd}). Upon observing their drawbacks of not being able to distill relational information, we further propose a {\color{mygreen} relational KD framework: \ours} (\cref{sec:llp}), which equips the student model with knowledge of each (anchor) node's relationships with other (context) nodes, via our proposed {\color{mygreen} rank-based matching} and {\color{mygreen} distribution-based matching} objectives.
    }
    \vspace{-0.1in}
    \label{fig:frame}
\end{figure*}

\textbf{Link Prediction with GNNs.} In this work, we take the commonly used encoder-decoder framework for the link prediction task~\citep{kipf2016variational,berg2017graph,schlichtkrull2018modeling,ying2018graph, davidson2018hyperspherical, zhu2021neural,yun2021neo,zhao2022learning}, where the GNN-based encoder learns node representations and the decoder predicts link existence probabilities.
Most GNNs operate under the message-passing framework, where
the model iteratively updates each node $i$'s representation $\boldsymbol{h}_i$ by fetching its neighbors' information. That is, the node's representation in the $l$-th layer is learned with an aggregation operation and an update operation:
\begin{equation}
\scale[0.94]{
    \boldsymbol{h}_i^l = \textsc{Update}_l \big(\boldsymbol{h}_i^{l-1}, \textsc{Aggregate}_l(\{\boldsymbol{h}_j^{l-1} | e_{i, j} \in \mathcal{E} \})\big),
}
\end{equation}
where $\textsc{Aggregate}$ combines or pools local neighbor features, $\textsc{\textsc{Update}}$ is a learnable transformation, and $\boldsymbol{h}_i^0 = \boldsymbol{x}_i$. 
The link prediction decoder takes the node representations from the last layer, i.e., $\boldsymbol{h}_i$ for $i \in \mathcal{V}$, to predict the probability of a link between a node pair $i$ and $j$:
\begin{equation}
    y_{i, j} = \sigma(\textsc{Decoder}(\boldsymbol{h}_i, \boldsymbol{h}_j)), 
    \label{equa: score}
\end{equation}
where $\sigma$ denotes a Sigmoid operation. In this work, following most state-of-the-art link prediction literature~\citep{zhang2021labeling,tsitsulin2018verse,zhao2022learning, wang2021pairwise}, we take the Hadamard product followed by a MLP as the link prediction $\textsc{Decoder}$ for all methods.

\textbf{Knowledge Distillation for GNNs.} Knowledge distillation (KD)~\citep{hinton2015distilling} aims to transfer the knowledge from a high-capacity and highly accurate teacher model to a light-weight student model, and is typically employed in resource-constrained settings. 
KD methods have shown great promise in significantly reducing model complexity, while sometimes barely (or not) sacrificing task performance \citep{furlanello2018born, park2019relational}. As GNNs are known to be slow due to neighbor aggregation induced by data dependency, graph-based KD methods~\citep{zhang2021graph,zheng2021cold} usually distill GNNs onto MLPs, which are commonly used in large-scale industrial applications due to their significantly improved efficiency and scalability. For example, \citet{zheng2021cold} proposed a KD-based framework to rediscover the missing graph structure information for MLPs, which improves the models' generalization of node classification task on cold-start nodes. Existing KD methods on graph data mainly focus on node-level \citep{zheng2021cold,zhang2021graph,tian2022nosmog} and graph-level tasks \citep{ma2019graph,zhang2020iterative,deng2021graph,joshi2021representation}, leaving KD for link prediction yet unexplored.
Our work focuses on distilling link prediction-relevant information from the GNN teacher to an MLP student, and investigates various KD strategies to align student and teacher predictions. 
Specifically, denoting the node representations for nodes $i$ and $j$ learned by the student MLP as $\hat{\boldsymbol{h}}_i$ and $\hat{\boldsymbol{h}}_j$, the link existence prediction by the student model can then be written as $\hat{y}_{i,j} = \sigma(\textsc{Decoder}(\hat{\boldsymbol{h}}_i, \hat{\boldsymbol{h}}_j))$.

\section{Cross-model Knowledge Distillation for Link Prediction}
In this section, we propose and discuss several approaches to distill knowledge from a teacher GNN to a student MLP in a cross-model fashion, for the purpose of link prediction. In all cases, we aim to supervise the student MLP with artifacts produced by the GNN teacher, in addition to any available training labels ($a_{i,j}$ w.l.o.g.) about link existence.  We start by adapting two \emph{direct} knowledge distillation (KD) methods: logit-matching and representation-matching, on link prediction tasks; we call these methods direct because they involve directly matching sample-wise predictions between teacher and student. Next, we motivate and introduce our proposed \emph{relational} KD framework, \ours, with two matching strategies to distill additional topology-related structural information to the student. We call these methods relational because they call for the preservation of relationships across samples between teacher and student \citep{park2019relational}. 
\cref{fig:frame} summarizes our proposals.

\vspace{-0.02in}
\subsection{Direct Methods}
\label{sec:directkd}
\textbf{Logit-matching} is one straightforward strategy to distill knowledge from the teacher to the student, where it directly aims to teach the student to generalize as the teacher does on the downstream task. It was proposed by \citet{hinton2015distilling}, and it is still one of the most widely used KD methods in various tasks~\citep{furlanello2018born, yang2020model, yan2020tinygnn}. Several works~\citep{phuong2019towards, ji2020knowledge} theoretically analyzed its effectiveness. Moreover, it had also been proved to be effective for knowledge transfer on graph data~\citep{yan2020tinygnn, yang2021extract, zhang2021graph} in recent years. For example, \citet{zhang2021graph} the soft logits generated by the teacher GNNs to help supervise the student MLP and achieved strong performance on node classification tasks. 
In a similar vein, we generate the soft score $y_{i, j}$ for the candidate edge ($i, j$) with the teacher GNN model, and train the student to match its prediction $\hat{y}_{i, j}$ on this target:
\begin{align}
\label{eq:lm}
    \mathcal{L}_{LM} =& \sum_{(i, j) \in \mathcal{E} \cup \mathcal{E}^{-}} \Big(\lambda\mathcal{L}_{sup} (\hat{y}_{i, j}, a_{i, j}) \nonumber\\
    &+ (1-\lambda) \mathcal{L}_{match}(\hat{y}_{i, j}, y_{i, j})\Big),
\end{align}
where $\mathcal{L}_{sup}$ is the supervised link prediction loss (e.g., binary cross entropy) that directly trains the student model, $\mathcal{L}_{match}$ is the loss for matching the student's prediction with the teacher's prediction, and $\lambda$ is a hyper-parameter that mediates the importance of the ground-truth labels and logit-matching signals. 
Note that multiple implementation choices exist for $\mathcal{L}_{match}$. For example, mean-squared error (MSE), Kullback-Leibler (KL) divergence, or cosine similarity. In the experiments, we opt for the empirical best choice for fair comparison across methods.

\textbf{Representation-matching} is another direct distillation method in which we aim to align the student's learned latent node embedding space with the teacher's. 
As this KD training signal only optimizes the encoder part of the student model, it must be used with $\mathcal{L}_{sup}$ so that the student decoder receives a gradient and can also be optimized:
\begin{align}
\label{eq:rm}
    \mathcal{L}_{RM} = &\sum_{(i, j) \in \mathcal{E} \cup \mathcal{E}^{-}} \lambda\mathcal{L}_{sup} (\hat{y}_{i, j}, a_{i, j}) \nonumber\\
    &+ (1-\lambda) \sum_{i \in \mathcal{V}} \mathcal{L}_{match}(\hat{\boldsymbol{h}}_i, \boldsymbol{h}_i).
\end{align}
Unlike logit-matching, representation-matching involves directly aligning node-level artifacts, which is similar to object representation matching in computer vision~\citep{romero2014fitnets, kim2018paraphrasing,wang2020exclusivity,chen2021cross}. 

\subsection{Link Prediction with Relational Distillation}
\label{sec:motivation}
\textbf{Motivation.} The above direct methods ask the student model to directly match node-level or link-level artifacts.  However, one might ask: are matching these sufficient for link prediction tasks? This is especially relevant considering that most link prediction applications involve tasks where ranking target nodes with respect to a source, or anchor node, is the task of interest, i.e. ranking relevant candidate users or items with respect to a seed user \citep{huang2005link, trouillon2016complex}. These contexts involve reasoning over multiple \emph{relations} or link-level samples simultaneously, suggesting that matching across these relations could be more aligned with the target link prediction task, compared to the direct node-level or link-level methods. 

Furthermore, several works \citep{zhang2018link, yun2021neo} suggest that graph structure information is critically important for link prediction tasks. For example, heuristic link prediction methods commonly show competitive performance compared to GNNs~\citep{zhang2018link} and have long-served as a cornerstone for accurate link prediction even prior to neural graph methods \citep{martinez2016survey}. Most heuristic methods measure the score of the target node pairs only based on the graph structure information~\citep{barabasi1999emergence, brin2012reprint}, such as common neighbors and shortest path. 
In addition, several recent works~\citep{zhang2017weisfeiler,zhang2018link,li2020distance,zhao2022learning} also show that enclosing topology information such as local subgraph, distances with anchor nodes, or augmented links can largely improve GNNs' performance on link-level tasks. Observing that most successful methods in link prediction involve using relational information other than just the two nodes in question, we also adopt this intuition in the distillation context and propose our relational KD for link prediction.  We elaborate next.

\subsection{Proposed Framework: \underline{L}inkless \underline{L}ink \underline{P}rediction}
\label{sec:llp}

In accordance with our intuition regarding preservation of relational knowledge, we propose a novel relational distillation framework, called \emph{\underline{L}inkless \underline{L}ink \underline{P}rediction}, or \ours. Instead of focusing on matching individual node pair scores or node representations, \ours focuses on distilling knowledge about the relationships of each node to other nodes in the graph; we call the former node an \emph{anchor} node, and the latter nodes \emph{context} nodes. For each node in the graph, when it serves as the anchor node, we aim to equip the student MLP model with knowledge of its relationships with a set of context nodes. Each node can serve as both an anchor node, as well as a context node (for other anchor nodes).

Let $v$ denote the anchor node 
and $\mathcal{C}_v$ denote the corresponding set of context nodes of $v$. We denote the teacher model's predicted probabilities of $v$ and each node in $\mathcal{C}_v$ as $\mathcal{Y}_v = \{y_{v,i} | i \in \mathcal{C}_v\}$.
Similarly, we denote the student model's predictions on those as $\hat{\mathcal{Y}}_v = \{\hat{y}_{v,i} | i \in \mathcal{C}_v\}$. To effectively distill the relational knowledge from $\mathcal{Y}_v$ to $\hat{\mathcal{Y}}_v$, we proposed two relational matching objectives to train \ours: \emph{rank-based matching} and \emph{distribution-based matching}, which we introduce next.

\textbf{Rank-based Matching.} 
As aforementioned in \cref{sec:motivation}, link prediction is often considered a ranking task, requiring the model to rank relevant candidates w.r.t. a seed node, e.g. in a user-item graph setting, the predictor must rank over a set of candidate items from the perspective of a user. Thus, we reason that unlike matching individual and  independent logits, matching the ranking induced by the teacher can more straightforwardly teach the student relational knowledge about context nodes w.r.t. the anchor node, e.g. for a specific user, item $A$ should be ranked higher than item $C$, which should be ranked higher than item $B$.  To adopt this rank-based intuition into a training objective, we adopt a modified margin-based ranking loss that trains the student with the rank of the logits from the teacher GNN. Specifically, we enumerate all pairs of predicted probabilities in $\hat{\mathcal{Y}}_v$ and supervise it with the corresponding pairs in $\mathcal{Y}_v$. That is, 
\begin{align}
\label{eq:loss_r}
    &\mathcal{L}_{\ours\_R} = \sum_{v \in \mathcal{V}} \enspace \sum_{\{\hat{y}_{v,i},\hat{y}_{v,j}\} \in \hat{\mathcal{Y}}_v} \max(0, -r \cdot (\hat{y}_{v, i} - \hat{y}_{v, j}) + \delta),\\
    &\text{where} \enspace r = \left\{
    \begin{aligned}1,\enspace &\text{if} \enspace y_{v, i} - y_{v, j} >  \delta; \\
    -1,\enspace &\text{if} \enspace y_{v, i} - y_{v, j} < - \delta; \\
    0,\enspace &\text{otherwise},
    \end{aligned} \right . \nonumber
\end{align}
where $\delta$ is the margin hyper-parameter,
which is usually a very small value (e.g. 0.05). Note that the above loss differs from the conventional margin-based ranking loss, because it has a condition for $r=0$ (inducing constant loss) on the logits pairs that the teacher GNN gives similar probabilities, i.e., $|y_{v, i} - y_{v, j}| < \delta$. This design effectively prevents the student model from trying to differentiate minuscule differences in probabilities which the teacher may produce  owing to noise; without this condition, the loss would pass binary information regardless of how small the difference is. We also empirically show the necessity of this design in~\cref{tab:delta} in~\cref{sec:apdx-results}.

\textbf{Distribution-based Matching.} While the rank-based matching can effectively teach the student model relational rank information, we observe that it does not fully make use of the value  information from $\mathcal{Y}_v$, e.g. for a specific user, item $A$ should be ranked \emph{much} higher than item $C$, which should only be ranked \emph{marginally} higher than item $B$. 
Although the logit-matching introduced in \cref{sec:directkd} might seem suitable here, we observe that its link-level matching strategy only facilitates matching information on scattered node pairs, rather than focusing on the relationships conditioned on an anchor node -- empirically, we also find that it has limited effectiveness.
Therefore, to enable relational value-based matching centered on the anchor nodes, we further propose a distribution-based matching scheme which utilizes the KL divergence between the teacher predictions $\mathcal{Y}_v$ and student predictions $\hat{\mathcal{Y}}_v$, centered on each anchor node $v$. Specifically, we define it as
\begin{align}
\label{eq:loss_d}
    \mathcal{L}_{\ours\_D} = \sum_{v \in \mathcal{V}} \enspace \sum_{i \in \mathcal{C}_v} &\frac{\exp(y_{v, i}/\tau)}{\sum_{j \in \mathcal{C}_v} \exp(y_{v, j}/\tau)} \nonumber\\ & \log \bigg(\frac{\exp(\hat{y}_{v, i}/\tau)}{\sum_{j \in \mathcal{C}_v} \exp(\hat{y}_{v, j}/\tau)} \bigg), 
\end{align}
where $\tau$ is a temperature hyper-parameter which controls the softness of the softmaxed distribution. 
By also asking the student to match relative values within the probability distribution over context nodes conditioned on each anchor node, the distribution-based matching scheme complements rank-based matching by providing auxiliary information about the magnitudes of differences. 

\textbf{Practical Implementation of \ours.} 
In practical implementation, given the large number of nodes in the graph, it is infeasible for \ours to use all other nodes as the set of context nodes, especially for the rank-based matching which enumerates pairs of probabilities in $\hat{\mathcal{Y}}_v$. Hence, we opt for simplicity and adopt two straightforward sampling strategies for the constructing $\mathcal{C}_v$ for each anchor node $v$ to limit its size. First, to keep the local structure around the anchor node, we follow previous works~\citep{perozzi2014deepwalk, hamilton2017inductive} to sample $p$ nearby nodes by repeating fixed-length random walks several times, denoted as $\mathcal{C}_v^N$. Secondly, we randomly sample $q$ nodes from the whole graph $G$ (which are likely to be far-away from $v$) to form $\mathcal{C}_v^R$, which additionally preserves the global structure w.r.t. $v$ in the graph. The context nodes for each anchor node are the union of the nearby samples and random samples. $p$ and $q$ are hyper-parameters. Finally, we make $\mathcal{C}_v$ as the union of the nearby samples and random samples, i.e., $\mathcal{C}_v = \mathcal{C}_v^N \cup \mathcal{C}_v^R$. We conduct experiments to show the impact of the selection strategy of context nodes $\mathcal{C}_v$ for each anchor node, which are presented in \cref{sec:ablation} and \cref{sec:apdx-context_node_selection}.

While training \ours, we jointly optimize both the rank-based and distribution-based matching losses in addition to the ground-truth label loss. Therefore, the overall training loss which \ours adopts for the student is
\begin{equation}
\label{eq:loss}
    \mathcal{L} = \alpha \cdot \mathcal{L}_{sup} + \beta \cdot \mathcal{L}_{\ours\_R} + \gamma \cdot \mathcal{L}_{\ours\_D}
\end{equation}
where $\alpha$, $\beta$, and $\gamma$ are hyper-parameters which mediate the strengths of each loss term.

\section{Experiments}
\label{sec:exp}
\subsection{Experimental Setup}
\textbf{Datasets.} We conduct the experiments using 8 commonly used benchmark datasets for link prediction: \cora, \citeseer, \pubmed, \computers, \photos, \cs, \physics, and \collab. The statistics of the datasets are shown in \cref{tab:statistic} with further details provided in \cref{sec:apdx-data}.
\begin{table}
\centering
\caption{Statistics of datasets.}
\label{tab:statistic}
\begin{tabular}{l|ccc} 
\toprule
Dataset    &\# Nodes    &\# Edges   &\# Features    \\ \midrule
\cora    &2,708 &5,278 &1,433\\
\citeseer &3,327 &4,552 &3,703\\
\pubmed &19,717 &44,324 &500\\
\cs &18,333 &163,788 &6,805\\   
\physics &34,493 &495,924 &8,415\\      
\computers &13,752 &491,722 &767\\    
\photos &7,650 &238,162 &745\\
\collab &235,868 &1,285,465 &128\\

\bottomrule
\end{tabular}
\end{table}

\textbf{Evaluation Settings.} 
To comprehensively evaluate our proposed \ours and baseline methods on the link prediction tasks, we conduct experiments on both transductive and production settings. For the transductive setting, all the nodes in the graph can be observed for train/validation/test sets. Following previous works~\citep{zhang2018link,chami2019hyperbolic,cai2021line} we randomly sample 5\%/15\% of the links with the same number of no-edge node pairs from the graph as the validation/test sets on the non-OGB datasets. And the validation/test links are masked off from the training graph. For the OGB datasets, we follow their official train/validation/test splits~\citep{wang2020microsoft}. In addition to transductive setting, we also design a more realistic setting that mimics practical link prediction use-cases, which we call the production setting. In the production setting, new nodes would appear in the test set, while training and validation sets only observe previously existing nodes. Thus, this setting entails three categories of node pairs (edges or no-edges) in the test set: existing -- existing, existing -- new, and new -- new, where the first category is similar to the test edges in the transductive setting, and the latter two categories together are similar to the inductive setting used in a few recent works~\citep{bojchevski2017deep, hao2020inductive, chen2021topology}. Nonetheless, all three types of these edges appear with varying proportions in practical use-cases, e.g. growth of a social network or online platform, hence we evaluate on all three types. Note that we only conduct production setting experiments on non-OGB datasets, because the OGB dataset is already temporally split in their public releases. We further elaborate the details of the production setting in \cref{sec:apdx-eval}.

For \collab, we use its official metric (Hits@50 for \collab) following their public leaderboard.
For other datasets, following previous works~\citep{yun2021neo,zhang2021labeling,zhao2022learning}, we use Hits@20 as the main metric, which is also one of the main metrics on OGB datasets. We also report AUC performance in \cref{sec:apdx-results}. For all experiments, we report the averaged test performance (with early-stopping on validation) along with its standard deviation over 10 runs with different random initializations.

\textbf{Reproducibility.} To ensure the reproducibility of \ours, our implementation is publically available at \url{https://github.com/snap-research/linkless-link-prediction/}.

\textbf{Methods.} In the remainder of this section: ``GNN'' indicates the teacher GNN that was trained with \lsup; ``MLP'' refers to the stand-alone MLP that was trained with \lsup; ``\kdp'' refers to MLP trained with logit matching (\cref{eq:lm}); ``\kdf'' refers to MLP trained with node representation matching (\cref{eq:rm}); ``\ours'' refers to MLP trained with our proposed relational KD (\cref{eq:loss}). For the main experiments, we opt for simplicity and use SAGE~\citep{hamilton2017inductive} as the teacher GNN in all settings. We also include further experiments of different teacher GNN models in \cref{apdx:diff_teacher,apdx:plnlp}.

\begin{table*}[t]
\centering
\caption{Link prediction performance under \textbf{transductive} setting. For \collab, we report Hits@50. For other datasets, we report Hits@20. Best and second best performances are marked with bold and underline, respectively. $\Delta_{DirectKD}$, $\Delta_{MLP}$, and $\Delta_{GNN}$ represent differences between \ours and these methods. }
\label{tab:trans_hits}
\begin{tabular}{l|cc|cc|cccccc} 
\toprule
    &GNN    &MLP    & \kdp   & \kdf   & \ours  &$\Delta_{DirectKD}$  & $\Delta_{MLP}$    & $\Delta_{GNN}$ \\ \midrule
\cora  &\ms{74.38} {1.54}  &\ms{\underline{78.06}} {1.50} &\ms{74.72} {4.27}   &\ms{75.75} {1.51}  &\ms{\textbf{78.82}} {1.74} & 3.07 & 0.76 & 4.44 \\

\citeseer  &\ms{\underline{73.89}} {0.95}  &\ms{71.21} {3.22}  &\ms{72.44} {1.52}  &\ms{65.19} {5.54}  &\ms{\textbf{77.32}} {2.42} & 4.88 &6.11 &3.43 \\

\pubmed  &\ms{\underline{51.98}} {5.25}  &\ms{42.89} {1.67}  &\ms{42.78} {3.15}  &\ms{44.44} {2.40}  &\ms{\textbf{57.33}} {2.42} & 12.89 & 14.44 & 5.35 \\

\cs  &\ms{59.51} {7.34}  &\ms{34.01} {9.37} &\ms{40.69} {5.12}  &\ms{\underline{61.10}} {2.83}  &\ms{\textbf{68.62}} {1.46} &7.52 &34.61 &9.11 \\

\physics  &\ms{\underline{66.74}} {1.53}  &\ms{31.26} {9.12} & \ms{52.11} {2.44} &\ms{52.34} {3.78} &\ms{\textbf{72.01}} {1.89} &19.67 & 40.75 &5.27 \\

\computers  &\ms{\underline{31.66}} {3.08} &\ms{20.19} {1.58}  &\ms{12.81} {1.80}  &\ms{21.75} {1.96}  &\ms{\textbf{35.32}} {2.28} &13.57 &15.31 &3.66 \\

\photos  &\ms{\textbf{51.50}} {4.48}  &\ms{27.83} {4.90}  &\ms{24.24} {2.79}  &\ms{38.47} {2.76}  &\ms{\underline{49.32}} {2.64} &10.85 &21.49 &-2.18 \\

\collab  & \ms{\underline{48.69}} {0.87} & \ms{36.95} {1.37} & \ms{35.97} {0.96} & \ms{36.86} {0.45} & \ms{\textbf{49.10}} {0.57} &12.24 &12.15 &0.41  \\


\bottomrule
\end{tabular}
\end{table*}

\subsection{Link Prediction Results}

\textbf{Transductive Setting.} \cref{tab:trans_hits} shows the link prediction performance of the proposed \ours with GNN, MLP, and the direct KD methods (as introduced in \cref{sec:directkd}) in the transductive setting. We observe that \ours consistently outperforms MLP and direct KD methods across all datasets with large margins. Specifically, \ours achieves \textbf{18.18} points and \textbf{10.59} points improvements over MLP and direct KD methods averaged on datasets, respectively. On the \physics dataset, \ours achieves \textbf{40.75} points and \textbf{19.67} points absolute improvements over MLP and direct KD, respectively. Moreover, \ours achieves better performance than the teacher GNN model on 7 out of 8 datasets, demonstrating that our proposed rank-based and distribution-based matching are able to effectively distill the knowledge for link prediction. 

From \cref{tab:trans_hits}, we also observe significant performance improvements of \ours over the teacher GNNs on some datasets. We hypothesize that there are two reasons leading to such improvements. The one is that the student MLP model already has significant learning ability when the node features are informative enough for link prediction. For example, on the \cora dataset, MLP already can produce better prediction performance than GNN. The other reason is relational KD provides relational structure knowledge from GNNs to MLPs, which provides extra valuable knowledge to MLPs. Recent studies on knowledge distillation~\citep{allen2020towards, guo2022boosting} also have similar findings that combining different views from different models could help improve the model's performance.

\begin{table*}[t]
\centering
\caption{Link prediction performance measured by Hits@20 under \textbf{production} setting. Best and second best performances are marked with bold and underline, respectively.}
\label{tab:product_hits}
\begin{tabular}{l|cc|cc|cccc}
\toprule
          & GNN     & MLP      & \kdp     & \kdf   & \ours &$\Delta_{DirectKD}$  & $\Delta_{MLP}$    & $\Delta_{GNN}$ \\
\midrule
\cora      & \ms{\underline{27.80}} {2.11}   & \ms{22.90} {2.22}  & \ms{22.65} {2.51} & \ms{22.24} {0.55} & \ms{\textbf{27.87}} {1.24} & 5.22  & 4.97  & 0.07   \\
\citeseer  & \ms{\textbf{38.78}} {2.59} & \ms{31.21} {3.75}  & \ms{29.35} {2.55} & \ms{26.23} {1.08} & \ms{\underline{34.75}} {2.45} & 5.40 & 3.54  & -4.03    \\
\pubmed    & \ms{\underline{52.71}} {1.81} & \ms{38.01} {1.67}  & \ms{39.03} {4.21} & \ms{43.27} {3.12} & \ms{\textbf{53.48}} {1.52} & 10.21 & 15.47 & 0.77    \\
\cs        & \ms{\underline{60.69}} {3.17} & \ms{38.15} {10.78} & \ms{48.07} {2.39} & \ms{58.90} {1.32} & \ms{\textbf{60.74}} {1.41} & 1.84 & 22.59 & 0.05    \\
\physics   & \ms{\textbf{55.82}} {2.43} & \ms{29.99} {1.96}  & \ms{22.74} {1.03} & \ms{36.32} {2.29} & \ms{\underline{52.83}} {1.50}  & 16.51 & 22.84 & -2.99   \\
\computers & \ms{\textbf{34.38}} {1.41} & \ms{19.43} {0.82}  & \ms{12.79} {1.43} & \ms{20.28} {1.01} & \ms{\underline{24.58}} {3.33} & 4.30 & 5.15  & -9.80      \\
\photos    & \ms{\textbf{51.03}} {6.05} & \ms{34.29} {2.49}  & \ms{24.63} {2.20}  & \ms{40.58} {1.63} & \ms{\underline{43.79}} {1.27} & 3.21 & 9.50  & -7.24   \\ 
\bottomrule
\end{tabular}
\vspace{-0.1in}
\end{table*}

\textbf{Production Setting.} \cref{tab:product_hits} shows the link prediction performance of the proposed \ours with GNN, MLP, and the direct KD methods in the production setting. From this table, we observe that \ours is still able to consistently outperform MLP and direct KD methods by large margins for all benchmarks. Specifically, \ours achieves \textbf{12.01} and \textbf{6.67} on Hits@20 improvements over MLP and direct KD methods averaged over datasets, respectively. Moreover, \ours is at or above par with the teacher GNN on 5 out of the 7 datasets, which supports deploying \ours is effective to distill the relational knowledge about link prediction from GNNs to MLPs. For ease of comparison, we also stratify each method's performance on the three different categories of the test edges in \cref{tab:product_hits_full} (in \cref{apdx:product_full}). 

In \cref{tab:product_hits}, we also observe that \ours achieves the in-stable performance across datasets, which is a recurring issue that plagues many related works~\citep{zhang2021graph, zheng2021cold} that study GNN to MLP distillation, especially under production or inductive settings. Under the transductive setting, \ours is able to impart a significant amount of relational knowledge to the student with respect to the nodes that already exist, and the evaluation will also be performed on those existing nodes. However, new nodes will emerge during the evaluation process under the production setting. In this setting, the efficacy of our approach depends on the quality of the original node features. The more informative the node features, the better our method will perform. This phenomenon is also aligned with Table 3 of GLNN~\citep{zhang2021graph}, which demonstrates a similar trend under the inductive setting for node classification tasks.

\begin{figure}[t]
    \centering
    {\includegraphics[width=0.49\textwidth]{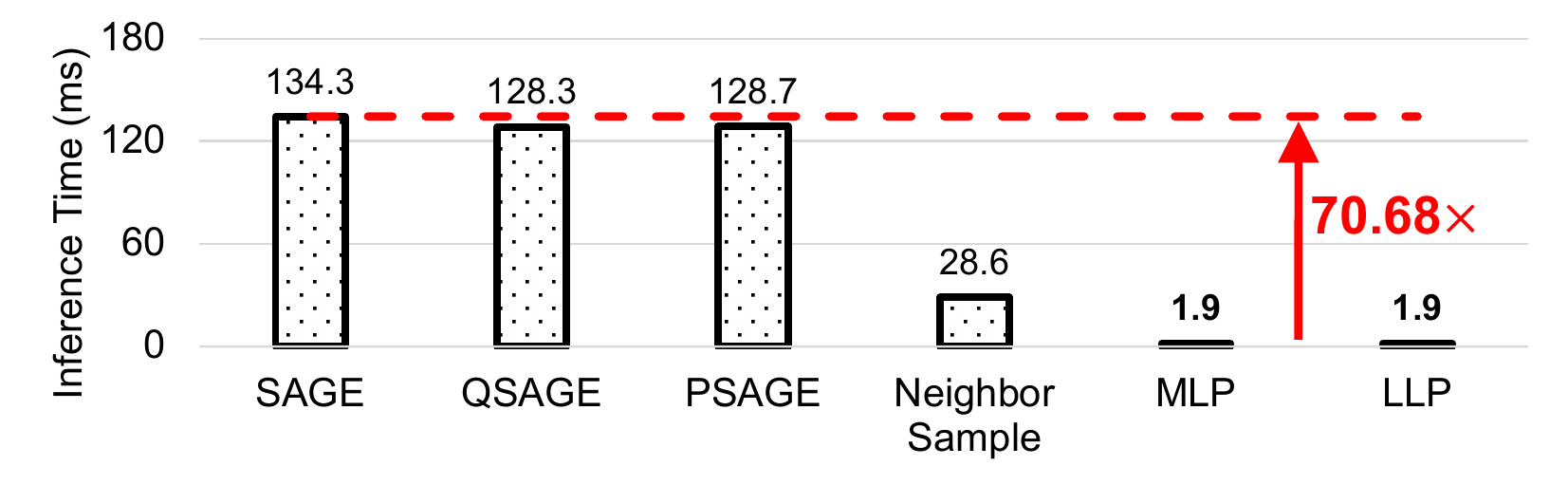}\label{fig:inference_collab}}
    {\includegraphics[width=0.49\textwidth]{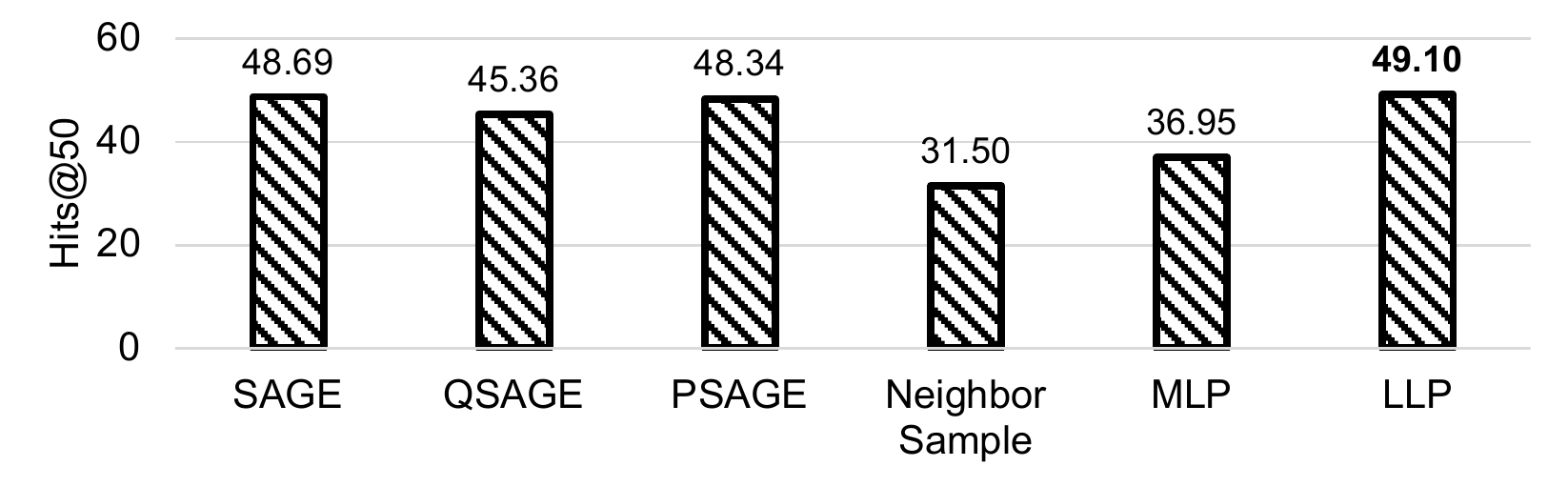}\label{fig:hits_collab}}
    \caption{Inference time and prediction performance comparison of \ours with GNN and GNN inference acceleration methods on \collab.}
    \label{fig:inference}
    \vspace{-0.15in}
\end{figure}

\begin{table}[t]
\caption{Link prediction performance measured by Hits@20 on \textbf{cold start nodes}.}
\label{tab:cold_start}
\resizebox{1\columnwidth}{!}{
\begin{tabular}{l|ccc|ccccc}
\toprule
& GNN   & MLP   & Ours  & $\Delta_{MLP}$ & $\Delta_{GNN}$   \\ 
\midrule
\cora      & 6.39  & 17.92 & \textbf{22.01} & 4.09 & 15.62   \\
\citeseer  & 11.04 & 29.33 & \textbf{32.09} & 2.76 & 21.05   \\
\pubmed    & 4.63  & 22.74 & \textbf{37.68} & 14.94 & 33.05  \\
\cs        & 9.46  & 29.09 & \textbf{46.83} & 17.74 & 37.37  \\
\physics   & 5.46  & 20.22 & \textbf{39.37} & 19.15 & 33.91  \\
\computers & 1.53  & 10.72 & \textbf{14.64} & 3.92 & 13.11   \\
\photos    & 0.87  & 20.44 & \textbf{23.79} & 3.35 & 22.92  \\
\bottomrule
\end{tabular}
}
\end{table}

\begin{table}[!ht]
\small
\centering
\caption{The statistics of large-scale datasets and the link prediction performances on them. ``\citationtwo-s'' indicates the down-sampled \citationtwo graph.}
\label{tab:largescale}
\scalebox{0.93}{
\begin{tabular}{l|cc|cc}
\toprule
& \igbtiny & \igbsmall & \citationtwo & \citationtwo-s \\
\midrule
\# Nodes & 100K  & 1M & 2,9M  & 122K   \\
\# Edges & 547K  & 12M & 30.6M & 1.4M \\
\midrule
GNN     & 79.25    & 40.12      & 82.56      & 39.99     \\
MLP     & 68.83    & 18.84      & 40.56      & 24.34     \\
\ours     & 79.47    & 39.78      & 53.20       & 29.23 \\
\bottomrule
\end{tabular}}
\end{table}

\subsection{Inference Acceleration Comparison}
We evaluate \ours in comparison to other common GNN inference acceleration methods, which mainly focus on the hardware and algorithm to reduce the computation consuming, such as pruning~\citep{zhou2021accelerating} and quantization~\citep{zhao2020learned, tailor2020degree}. Following the experimental settings in ~\citet{zhang2021graph}, we measure the inductive inference time on in the graph. We evaluate against 4 common GNN inference acceleration methods: (i) SAGE~\citep{hamilton2017inductive}, (ii) Quantized SAGE (QSAGE)~\citep{zhao2020learned, tailor2020degree} from \verb+float32+ to \verb+int8+, (iii) SAGE with 50\% weights pruned (PSAGE)~\citep{zhou2021accelerating,chen2021unified}, and (iv) SAGE with Neighbor Sampling with fan-out 15. 
\cref{fig:inference} shows the results on the large-scale OGB dataset, \collab. We can observe that \ours can obtain \textbf{70.68$\times$} speedup comparing with on SAGE on \collab. Compared with the best acceleration method Neighbor Sampling (which reduces graph dependency, but does not eliminate it like \ours), \ours still achieves \textbf{15.05$\times$} speedup. 
This is because all these inference acceleration methods still rely on the graph structure. From the bottom figure in~\cref{fig:inference}, we can further observe that \ours can outperform both GNN and other inference acceleration methods. 

\subsection{Link Prediction Results on Cold Start Nodes}
\label{sec:cold}

Dealing with cold start nodes (newly appeared nodes without edges) is a common challenge in recommendation and information retrieval applications~\citep{li2019zero, zheng2021cold, ding2021zero}. Without these edges, GNNs cannot perform well as they rely heavily on neighbor information.  On the other hand, MLPs, which do not make use of any graph topology information, are arguably more suitable. Here, we simulate the cold-start setting by removing all the new edges during testing stage of the production setting, i.e. all the new appeared nodes are isolated (more details are shown in \cref{sec:apdx-eval}). \cref{tab:cold_start} shows the performances of \ours, the stand-alone MLP, and the teacher GNN on the cold-start nodes. We observe that \ours consistently outperforms GNN and MLP by average of \textbf{25.29} and \textbf{9.42} on Hits@20, respectively. We further compare \ours with another related work on cold-start nodes in \cref{sec:cold-start}

\subsection{Link Prediction Results on Large Scale Datasets} 
Besides \collab, we also conduct experiments on three recently proposed large-scale graph benchmarks, \igbtiny, \igbsmall~\citep{igbdatasets}, and \citationtwo~\citep{wang2020microsoft, mikolov2013distributed}. The dataset stats and link prediction results (Hits@200) are shown in~\cref{tab:largescale}. On the IGB datasets, we observe that \ours can produce competitive results on these two datasets, which further demonstrates \ours has the ability to acquire complex link prediction-related knowledge from large-scale graphs. On the other hand, the results on \citationtwo show different patterns. Although \ours is able to significantly outperform MLP, its performances still show big gaps when compared with GNNs. 

We hypothesize that the different performances pattern on \citationtwo is due to the dataset's unique distribution. To validate our hypothesis, we further conduct experiments on a sampled version of \citationtwo (the ``\citationtwo-s'' column in \cref{tab:largescale}). Specifically, we down-sample \citationtwo to produce a smaller version of it (with size similar to \collab). We use random walk-based sampling, which has proved ability of property preserving on the original graph ~\citep{leskovec2006sampling}. From \cref{tab:largescale} we observe very similar patterns on the performances on both \citationtwo and \citationtwo-s, validation our hypothesize that the performance gap between \ours and GNNs are due to the dataset's own distribution rather than other factors such as the size of the graph.
Similar observation can also be found when comparing the results of \photos and \collab in \cref{tab:trans_hits}, but in a reversed way: \ours performs better than GNNs on the larger dataset (\collab) but slightly worse on the smaller one (\photos). In summary, the effectiveness of \ours may vary on different datasets, but is not sensitive to the size of the graphs.

\subsection{Ablation Study}
\label{sec:ablation}
\begin{table}[t]
\caption{Link prediction performances measured by Hits@20 on different components of \ours.}
\label{tab:ablation_rd}
\resizebox{1\columnwidth}{!}{
\begin{tabular}{l|cc|cc}
\toprule
Setting & \multicolumn{2}{c|}{Transductive} & \multicolumn{2}{c}{Production} \\
Dataset & \pubmed & \cs & \pubmed & \cs \\
\midrule
GNN & 51.98 & 59.51 & 52.71 & 60.69 \\
MLP & 42.89 & 40.69 & 38.01 & 38.15 \\
\midrule
\midrule
\ours & \textbf{57.33} & \textbf{68.62} & \textbf{53.48} & \textbf{60.74} \\
w/o \kdr & 55.35 & 66.61 & 53.40 & 60.53 \\
w/o \kdkl & 54.97 & 65.17 & 48.58 & 60.13 \\
w/o \kdr, \lsup & 54.86 & 68.39 & 39.35 & 57.35 \\
w/o \kdkl, \lsup & 53.30 & 68.30 & 41.43 & 55.63 \\
\bottomrule
\end{tabular}
}
\end{table}
\textbf{Effectiveness of \kdr and \kdkl.} As our proposed \ours contains two matching strategies, rank-based and distribution-based matching, we evaluate their effectiveness by removing them from \ours. Moreover, we further evaluate by also removing \lsup, i.e., using only one of the matching losses as the overall loss for \ours. \cref{tab:ablation_rd} shows the results of these settings compared with the performances of full \ours, stand-alone MLP, and the teacher GNN on \pubmed and \cs datasets under both settings.
We observe that both rank-based and distribution-based matching contribute significantly for the overall performance. In the transductive setting, both loss terms by themselves (the bottom two rows) can already outperform the teacher GNN. In the production setting, the matching losses alone outperform MLP and can achieve comparable performances with GNN after \lsup is added. In conclusion, both rank-based and distribution-based matching can effectively distill the relational knowledge, and they achieve the best performance by complementing each other.

\begin{figure}[t]
    \centering
    {\includegraphics[width=0.48\textwidth]{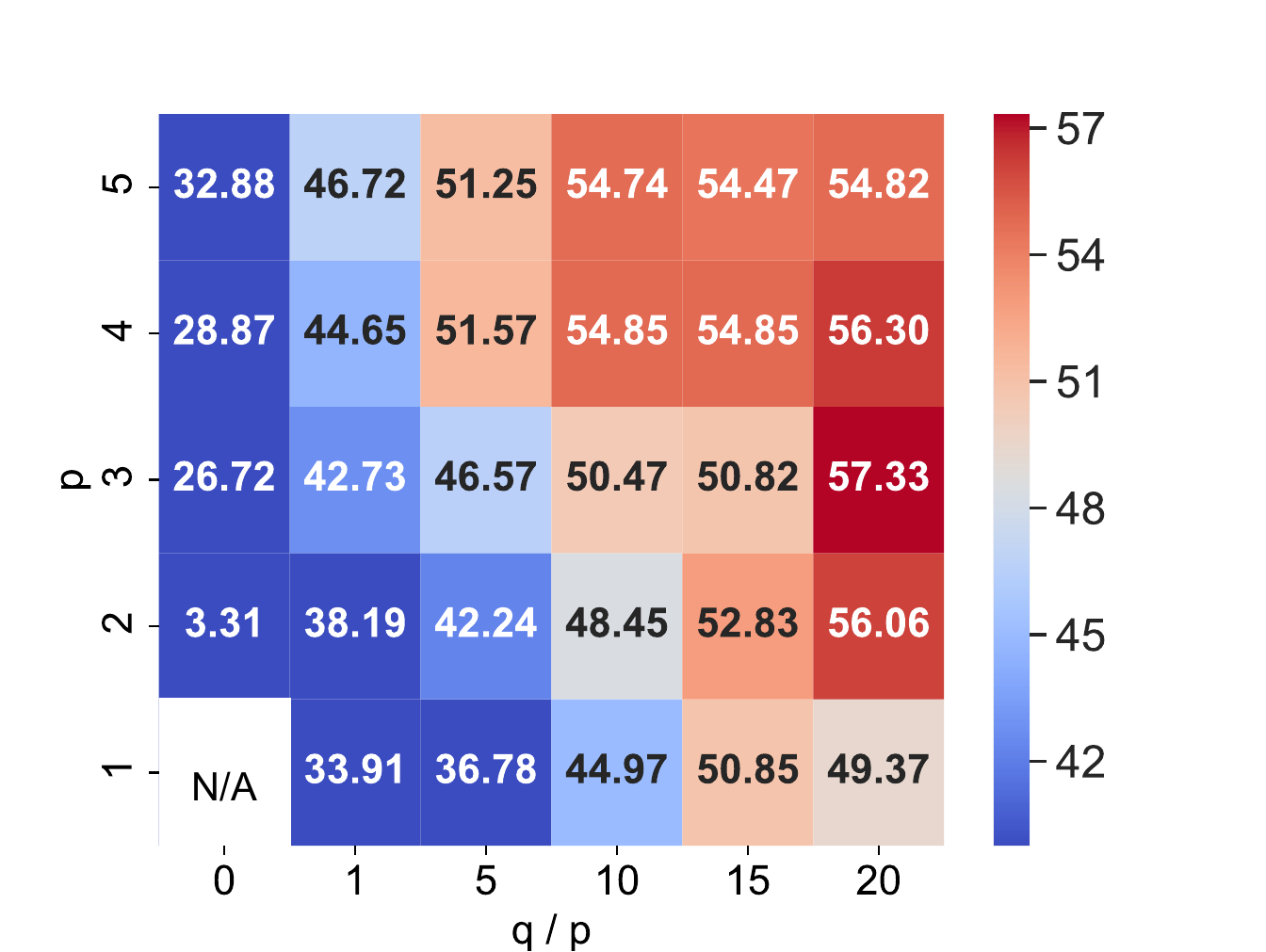}}
    \caption{Link prediction performance measured by Hits@20 of \ours on \pubmed with different number of samples for the context nodes.}
    \label{fig:sensitive}
\end{figure}

\textbf{Context Sampling Sensitivity to $p$ and $q$.} 
\cref{fig:sensitive} shows the link prediction performance of \ours on \pubmed under the transductive setting with different numbers of  context node samples ($p$ local samples, and $q$ random samples). For the ease of hyper-parameter tuning, we make $q$ a multiple of $p$, as shown in the $x$-axis of \cref{fig:sensitive}. We observe that \ours with low number of random samples shows poor link prediction performances, suggesting that preserving global relations are necessary for the proposed relation KD. Generally, the heatmap shows a clear trend, making the optimal values easy to locate.

\section{Conclusion}
Our work tackled problems related to applying GNNs for link prediction at scale.  We note these models have high latency at inference time owing to non-trivial data dependency.  In response, we explored applying cross-model distillation methods from teacher GNN to student MLP models, which are advantaged in inference time.  We first adopt two direct logit matching and representation matching KD methods to the link prediction context and observe their unsuitability.  In response, we introduced a relational KD framework, \ours, which proposed \emph{rank-based matching} and \emph{distribution-based matching} objectives which complement each other to force the student to preserve key information about contextual relationships across anchor nodes.  Our experiments demonstrated that \ours achieved MLP-level speedups (up to \textbf{70.68}\texttimes\xspace over GNNs), while also improving link prediction performance over MLPs by \textbf{18.18} and \textbf{12.01} points in transductive and production settings, matching or outperforming the teacher GNN in \textbf{7/8} datasets in transductive setting and 3/6 datasets in production setting, and with notable \textbf{25.29} on Hits@20 improvements on cold-start nodes.

\section*{Limitations} Similar to other MLP-based graph learning methods~\citep{zhang2021graph, zheng2021cold}, the main limitation of our proposed \ours is that it relies heavily on feature quality. It can hardly perform well when the node features are unreliable or missing.

\section*{Ethical Impact}
We do not foresee any negative societal impact or ethical concerns posed by our method. Nonetheless, we note that both positive and negative societal impacts can be made by applications of graph machine learning techniques, which may benefit from the improvements induced by our work. Care must be taken, in general, to ensure positive societal and ethical consequences of machine learning.

\section*{Acknowledgement}
We appreciate Xiaotian Han from Texas A\&M University, Wei Jin from Michigan State University, and Yiwei Wang from National University of Singapore for valuable discussions and suggestions. 

\bibliography{ref}

\begin{thebibliography}{112}
\providecommand{\natexlab}[1]{#1}
\providecommand{\url}[1]{\texttt{#1}}
\expandafter\ifx\csname urlstyle\endcsname\relax
  \providecommand{\doi}[1]{doi: #1}\else
  \providecommand{\doi}{doi: \begingroup \urlstyle{rm}\Url}\fi

\bibitem[Adamic \& Adar(2003)Adamic and Adar]{adamic2003friends}
Adamic, L.~A. and Adar, E.
\newblock Friends and neighbors on the web.
\newblock \emph{Social networks}, 2003.

\bibitem[Allen-Zhu \& Li(2020)Allen-Zhu and Li]{allen2020towards}
Allen-Zhu, Z. and Li, Y.
\newblock Towards understanding ensemble, knowledge distillation and
  self-distillation in deep learning.
\newblock \emph{arXiv preprint arXiv:2012.09816}, 2020.

\bibitem[Alon \& Yahav(2020)Alon and Yahav]{alon2020bottleneck}
Alon, U. and Yahav, E.
\newblock On the bottleneck of graph neural networks and its practical
  implications.
\newblock \emph{arXiv preprint arXiv:2006.05205}, 2020.

\bibitem[Barab{\'a}si \& Albert(1999)Barab{\'a}si and
  Albert]{barabasi1999emergence}
Barab{\'a}si, A.-L. and Albert, R.
\newblock Emergence of scaling in random networks.
\newblock \emph{science}, 1999.

\bibitem[Berg et~al.(2017)Berg, Kipf, and Welling]{berg2017graph}
Berg, R. v.~d., Kipf, T.~N., and Welling, M.
\newblock Graph convolutional matrix completion.
\newblock \emph{arXiv preprint arXiv:1706.02263}, 2017.

\bibitem[Bevilacqua et~al.(2021)Bevilacqua, Frasca, Lim, Srinivasan, Cai,
  Balamurugan, Bronstein, and Maron]{bevilacqua2021equivariant}
Bevilacqua, B., Frasca, F., Lim, D., Srinivasan, B., Cai, C., Balamurugan, G.,
  Bronstein, M.~M., and Maron, H.
\newblock Equivariant subgraph aggregation networks.
\newblock \emph{arXiv preprint arXiv:2110.02910}, 2021.

\bibitem[Bojchevski \& G{\"u}nnemann(2017)Bojchevski and
  G{\"u}nnemann]{bojchevski2017deep}
Bojchevski, A. and G{\"u}nnemann, S.
\newblock Deep gaussian embedding of graphs: Unsupervised inductive learning
  via ranking.
\newblock \emph{arXiv preprint arXiv:1707.03815}, 2017.

\bibitem[Brin \& Page(2012)Brin and Page]{brin2012reprint}
Brin, S. and Page, L.
\newblock Reprint of: The anatomy of a large-scale hypertextual web search
  engine.
\newblock \emph{Computer networks}, 2012.

\bibitem[Cai \& Ji(2020)Cai and Ji]{cai2020multi}
Cai, L. and Ji, S.
\newblock A multi-scale approach for graph link prediction.
\newblock In \emph{Proceedings of the AAAI conference on artificial
  intelligence}, 2020.

\bibitem[Cai et~al.(2021)Cai, Li, Wang, and Ji]{cai2021line}
Cai, L., Li, J., Wang, J., and Ji, S.
\newblock Line graph neural networks for link prediction.
\newblock \emph{IEEE Transactions on Pattern Analysis and Machine
  Intelligence}, 2021.

\bibitem[Cao et~al.(2007)Cao, Qin, Liu, Tsai, and Li]{cao2007learning}
Cao, Z., Qin, T., Liu, T.-Y., Tsai, M.-F., and Li, H.
\newblock Learning to rank: from pairwise approach to listwise approach.
\newblock In \emph{Proceedings of the 24th international conference on Machine
  learning}, pp.\  129--136, 2007.

\bibitem[Chami et~al.(2019)Chami, Ying, R{\'e}, and
  Leskovec]{chami2019hyperbolic}
Chami, I., Ying, Z., R{\'e}, C., and Leskovec, J.
\newblock Hyperbolic graph convolutional neural networks.
\newblock \emph{Advances in neural information processing systems}, 2019.

\bibitem[Chen et~al.(2021{\natexlab{a}})Chen, Mei, Zhang, Wang, Wang, Feng, and
  Chen]{chen2021cross}
Chen, D., Mei, J.-P., Zhang, Y., Wang, C., Wang, Z., Feng, Y., and Chen, C.
\newblock Cross-layer distillation with semantic calibration.
\newblock In \emph{Proceedings of the AAAI Conference on Artificial
  Intelligence}, 2021{\natexlab{a}}.

\bibitem[Chen et~al.(2021{\natexlab{b}})Chen, He, Wu, and
  Wang]{chen2021topology}
Chen, J., He, H., Wu, F., and Wang, J.
\newblock Topology-aware correlations between relations for inductive link
  prediction in knowledge graphs.
\newblock In \emph{Proceedings of the AAAI Conference on Artificial
  Intelligence}, 2021{\natexlab{b}}.

\bibitem[Chen et~al.(2020)Chen, Wei, Huang, Ding, and Li]{chen2020simple}
Chen, M., Wei, Z., Huang, Z., Ding, B., and Li, Y.
\newblock Simple and deep graph convolutional networks.
\newblock In \emph{International Conference on Machine Learning}, 2020.

\bibitem[Chen et~al.(2021{\natexlab{c}})Chen, Sui, Chen, Zhang, and
  Wang]{chen2021unified}
Chen, T., Sui, Y., Chen, X., Zhang, A., and Wang, Z.
\newblock A unified lottery ticket hypothesis for graph neural networks.
\newblock In \emph{International Conference on Machine Learning}, pp.\
  1695--1706. PMLR, 2021{\natexlab{c}}.

\bibitem[Covington et~al.(2016)Covington, Adams, and Sargin]{covington2016deep}
Covington, P., Adams, J., and Sargin, E.
\newblock Deep neural networks for youtube recommendations.
\newblock In \emph{Proceedings of the 10th ACM conference on recommender
  systems}, pp.\  191--198, 2016.

\bibitem[Davidson et~al.(2018)Davidson, Falorsi, De~Cao, Kipf, and
  Tomczak]{davidson2018hyperspherical}
Davidson, T.~R., Falorsi, L., De~Cao, N., Kipf, T., and Tomczak, J.~M.
\newblock Hyperspherical variational auto-encoders.
\newblock \emph{arXiv preprint arXiv:1804.00891}, 2018.

\bibitem[Deng \& Zhang(2021)Deng and Zhang]{deng2021graph}
Deng, X. and Zhang, Z.
\newblock Graph-free knowledge distillation for graph neural networks.
\newblock \emph{arXiv preprint arXiv:2105.07519}, 2021.

\bibitem[Ding et~al.(2021)Ding, Ma, Deoras, Wang, and Wang]{ding2021zero}
Ding, H., Ma, Y., Deoras, A., Wang, Y., and Wang, H.
\newblock Zero-shot recommender systems.
\newblock \emph{arXiv preprint arXiv:2105.08318}, 2021.

\bibitem[Fan et~al.(2022)Fan, Liu, Jin, Zhao, Tang, and Li]{fan2022graph}
Fan, W., Liu, X., Jin, W., Zhao, X., Tang, J., and Li, Q.
\newblock Graph trend filtering networks for recommendation.
\newblock In \emph{Proceedings of the 45th International ACM SIGIR Conference
  on Research and Development in Information Retrieval}, pp.\  112--121, 2022.

\bibitem[Fey \& Lenssen(2019)Fey and Lenssen]{fey2019fast}
Fey, M. and Lenssen, J.~E.
\newblock Fast graph representation learning with pytorch geometric.
\newblock \emph{arXiv preprint arXiv:1903.02428}, 2019.

\bibitem[Fey et~al.(2021)Fey, Lenssen, Weichert, and
  Leskovec]{fey2021gnnautoscale}
Fey, M., Lenssen, J.~E., Weichert, F., and Leskovec, J.
\newblock Gnnautoscale: Scalable and expressive graph neural networks via
  historical embeddings.
\newblock In \emph{International Conference on Machine Learning}, pp.\
  3294--3304. PMLR, 2021.

\bibitem[Furlanello et~al.(2018)Furlanello, Lipton, Tschannen, Itti, and
  Anandkumar]{furlanello2018born}
Furlanello, T., Lipton, Z., Tschannen, M., Itti, L., and Anandkumar, A.
\newblock Born again neural networks.
\newblock In \emph{International Conference on Machine Learning}, pp.\
  1607--1616. PMLR, 2018.

\bibitem[Geerts \& Reutter(2022)Geerts and Reutter]{geerts2022expressiveness}
Geerts, F. and Reutter, J.~L.
\newblock Expressiveness and approximation properties of graph neural networks.
\newblock \emph{arXiv preprint arXiv:2204.04661}, 2022.

\bibitem[Gholami et~al.(2021)Gholami, Kim, Dong, Yao, Mahoney, and
  Keutzer]{gholami2021survey}
Gholami, A., Kim, S., Dong, Z., Yao, Z., Mahoney, M.~W., and Keutzer, K.
\newblock A survey of quantization methods for efficient neural network
  inference.
\newblock \emph{arXiv preprint arXiv:2103.13630}, 2021.

\bibitem[Gilmer et~al.(2017)Gilmer, Schoenholz, Riley, Vinyals, and
  Dahl]{gilmer2017neural}
Gilmer, J., Schoenholz, S.~S., Riley, P.~F., Vinyals, O., and Dahl, G.~E.
\newblock Neural message passing for quantum chemistry.
\newblock In \emph{International conference on machine learning}, pp.\
  1263--1272. PMLR, 2017.

\bibitem[Gou et~al.(2021)Gou, Yu, Maybank, and Tao]{gou2021knowledge}
Gou, J., Yu, B., Maybank, S.~J., and Tao, D.
\newblock Knowledge distillation: A survey.
\newblock \emph{International Journal of Computer Vision}, 2021.

\bibitem[Guo et~al.(2021)Guo, Zhang, Yu, Herr, Wiest, Jiang, and
  Chawla]{guo2021few}
Guo, Z., Zhang, C., Yu, W., Herr, J., Wiest, O., Jiang, M., and Chawla, N.~V.
\newblock Few-shot graph learning for molecular property prediction.
\newblock In \emph{WWW}, 2021.

\bibitem[Guo et~al.(2022)Guo, Zhang, Fan, Tian, Zhang, and
  Chawla]{guo2022boosting}
Guo, Z., Zhang, C., Fan, Y., Tian, Y., Zhang, C., and Chawla, N.
\newblock Boosting graph neural networks via adaptive knowledge distillation.
\newblock \emph{arXiv preprint arXiv:2210.05920}, 2022.

\bibitem[Hamilton et~al.(2017)Hamilton, Ying, and
  Leskovec]{hamilton2017inductive}
Hamilton, W., Ying, Z., and Leskovec, J.
\newblock Inductive representation learning on large graphs.
\newblock \emph{Advances in neural information processing systems}, 2017.

\bibitem[Hao et~al.(2020)Hao, Cao, Fang, Xie, and Wang]{hao2020inductive}
Hao, Y., Cao, X., Fang, Y., Xie, X., and Wang, S.
\newblock Inductive link prediction for nodes having only attribute
  information.
\newblock \emph{arXiv preprint arXiv:2007.08053}, 2020.

\bibitem[He et~al.(2020)He, Deng, Wang, Li, Zhang, and Wang]{he2020lightgcn}
He, X., Deng, K., Wang, X., Li, Y., Zhang, Y., and Wang, M.
\newblock Lightgcn: Simplifying and powering graph convolution network for
  recommendation.
\newblock In \emph{Proceedings of the 43rd International ACM SIGIR conference
  on research and development in Information Retrieval}, pp.\  639--648, 2020.

\bibitem[Hinton et~al.(2015)Hinton, Vinyals, Dean,
  et~al.]{hinton2015distilling}
Hinton, G., Vinyals, O., Dean, J., et~al.
\newblock Distilling the knowledge in a neural network.
\newblock \emph{arXiv preprint arXiv:1503.02531}, 2015.

\bibitem[Hu et~al.(2021)Hu, You, Wang, Wang, Zhou, and Gao]{hu2021graph}
Hu, Y., You, H., Wang, Z., Wang, Z., Zhou, E., and Gao, Y.
\newblock Graph-mlp: node classification without message passing in graph.
\newblock \emph{arXiv preprint arXiv:2106.04051}, 2021.

\bibitem[Huang et~al.(2005)Huang, Li, and Chen]{huang2005link}
Huang, Z., Li, X., and Chen, H.
\newblock Link prediction approach to collaborative filtering.
\newblock In \emph{Proceedings of the 5th ACM/IEEE-CS joint conference on
  Digital libraries}, pp.\  141--142, 2005.

\bibitem[Jeh \& Widom(2002)Jeh and Widom]{jeh2002simrank}
Jeh, G. and Widom, J.
\newblock Simrank: a measure of structural-context similarity.
\newblock In \emph{Proceedings of the eighth ACM SIGKDD international
  conference on Knowledge discovery and data mining}, 2002.

\bibitem[Ji \& Zhu(2020)Ji and Zhu]{ji2020knowledge}
Ji, G. and Zhu, Z.
\newblock Knowledge distillation in wide neural networks: Risk bound, data
  efficiency and imperfect teacher.
\newblock \emph{Advances in Neural Information Processing Systems}, 2020.

\bibitem[Jia et~al.(2020)Jia, Lin, Ying, You, Leskovec, and
  Aiken]{jia2020redundancy}
Jia, Z., Lin, S., Ying, R., You, J., Leskovec, J., and Aiken, A.
\newblock Redundancy-free computation for graph neural networks.
\newblock In \emph{Proceedings of the 26th ACM SIGKDD International Conference
  on Knowledge Discovery \& Data Mining}, 2020.

\bibitem[Joshi et~al.(2021)Joshi, Liu, Xun, Lin, and
  Foo]{joshi2021representation}
Joshi, C.~K., Liu, F., Xun, X., Lin, J., and Foo, C.-S.
\newblock On representation knowledge distillation for graph neural networks.
\newblock \emph{arXiv preprint arXiv:2111.04964}, 2021.

\bibitem[Ju et~al.(2023)Ju, Zhao, Wen, Yu, Shah, Ye, and Zhang]{ju2023multi}
Ju, M., Zhao, T., Wen, Q., Yu, W., Shah, N., Ye, Y., and Zhang, C.
\newblock Multi-task self-supervised graph neural networks enable stronger task
  generalization.
\newblock \emph{International Conference on Learning Representations}, 2023.

\bibitem[Kang et~al.(2021)Kang, Hwang, Kweon, and Yu]{kang2021topology}
Kang, S., Hwang, J., Kweon, W., and Yu, H.
\newblock Topology distillation for recommender system.
\newblock In \emph{Proceedings of the 27th ACM SIGKDD Conference on Knowledge
  Discovery \& Data Mining}, pp.\  829--839, 2021.

\bibitem[Khatua et~al.(2023)Khatua, Mailthody, Taleka, Ma, Song, and
  Hwu]{igbdatasets}
Khatua, A., Mailthody, V.~S., Taleka, B., Ma, T., Song, X., and Hwu, W.-m.
\newblock Igb: Addressing the gaps in labeling, features, heterogeneity, and
  size of public graph datasets for deep learning research.
\newblock In \emph{In Proceedings of the 29th ACM SIGKDD Conference on
  Knowledge Discovery and Data Mining}, 2023.

\bibitem[Kim et~al.(2018)Kim, Park, and Kwak]{kim2018paraphrasing}
Kim, J., Park, S., and Kwak, N.
\newblock Paraphrasing complex network: Network compression via factor
  transfer.
\newblock \emph{Advances in neural information processing systems}, 2018.

\bibitem[Kipf \& Welling(2016{\natexlab{a}})Kipf and Welling]{kipf2016semi}
Kipf, T.~N. and Welling, M.
\newblock Semi-supervised classification with graph convolutional networks.
\newblock \emph{arXiv preprint arXiv:1609.02907}, 2016{\natexlab{a}}.

\bibitem[Kipf \& Welling(2016{\natexlab{b}})Kipf and
  Welling]{kipf2016variational}
Kipf, T.~N. and Welling, M.
\newblock Variational graph auto-encoders.
\newblock \emph{arXiv preprint arXiv:1611.07308}, 2016{\natexlab{b}}.

\bibitem[Koren et~al.(2009)Koren, Bell, and Volinsky]{koren2009matrix}
Koren, Y., Bell, R., and Volinsky, C.
\newblock Matrix factorization techniques for recommender systems.
\newblock \emph{Computer}, 2009.

\bibitem[Leskovec \& Faloutsos(2006)Leskovec and
  Faloutsos]{leskovec2006sampling}
Leskovec, J. and Faloutsos, C.
\newblock Sampling from large graphs.
\newblock In \emph{Proceedings of the 12th ACM SIGKDD international conference
  on Knowledge discovery and data mining}, 2006.

\bibitem[Li et~al.(2019)Li, Jing, Lu, Zhu, Yang, and Huang]{li2019zero}
Li, J., Jing, M., Lu, K., Zhu, L., Yang, Y., and Huang, Z.
\newblock From zero-shot learning to cold-start recommendation.
\newblock In \emph{Proceedings of the AAAI conference on artificial
  intelligence}, 2019.

\bibitem[Li et~al.(2020)Li, Wang, Wang, and Leskovec]{li2020distance}
Li, P., Wang, Y., Wang, H., and Leskovec, J.
\newblock Distance encoding: Design provably more powerful neural networks for
  graph representation learning.
\newblock \emph{Advances in Neural Information Processing Systems},
  33:\penalty0 4465--4478, 2020.

\bibitem[Liu et~al.(2022)Liu, Zhao, Xu, Luo, and Jiang]{liu2022graph}
Liu, G., Zhao, T., Xu, J., Luo, T., and Jiang, M.
\newblock Graph rationalization with environment-based augmentations.
\newblock In \emph{Proceedings of the 28th ACM SIGKDD International Conference
  on Knowledge Discovery \& Data Mining}, 2022.

\bibitem[Liu et~al.(2021)Liu, Jin, Ma, Li, Liu, Wang, Yan, and
  Tang]{liu2021elastic}
Liu, X., Jin, W., Ma, Y., Li, Y., Liu, H., Wang, Y., Yan, M., and Tang, J.
\newblock Elastic graph neural networks.
\newblock In \emph{International Conference on Machine Learning}, pp.\
  6837--6849. PMLR, 2021.

\bibitem[Ma \& Mei(2019)Ma and Mei]{ma2019graph}
Ma, J. and Mei, Q.
\newblock Graph representation learning via multi-task knowledge distillation.
\newblock \emph{arXiv preprint arXiv:1911.05700}, 2019.

\bibitem[Ma et~al.(2021)Ma, Liu, Zhao, Liu, Tang, and Shah]{ma2021unified}
Ma, Y., Liu, X., Zhao, T., Liu, Y., Tang, J., and Shah, N.
\newblock A unified view on graph neural networks as graph signal denoising.
\newblock In \emph{Proceedings of the 30th ACM International Conference on
  Information \& Knowledge Management}, 2021.

\bibitem[Maron et~al.(2019)Maron, Ben-Hamu, Serviansky, and
  Lipman]{maron2019provably}
Maron, H., Ben-Hamu, H., Serviansky, H., and Lipman, Y.
\newblock Provably powerful graph networks.
\newblock \emph{Advances in neural information processing systems}, 32, 2019.

\bibitem[Mart{\'\i}nez et~al.(2016)Mart{\'\i}nez, Berzal, and
  Cubero]{martinez2016survey}
Mart{\'\i}nez, V., Berzal, F., and Cubero, J.-C.
\newblock A survey of link prediction in complex networks.
\newblock \emph{ACM computing surveys (CSUR)}, 49\penalty0 (4):\penalty0 1--33,
  2016.

\bibitem[McAuley et~al.(2015)McAuley, Targett, Shi, and Van
  Den~Hengel]{mcauley2015image}
McAuley, J., Targett, C., Shi, Q., and Van Den~Hengel, A.
\newblock Image-based recommendations on styles and substitutes.
\newblock In \emph{Proceedings of the 38th international ACM SIGIR conference
  on research and development in information retrieval}, 2015.

\bibitem[Mikolov et~al.(2013)Mikolov, Sutskever, Chen, Corrado, and
  Dean]{mikolov2013distributed}
Mikolov, T., Sutskever, I., Chen, K., Corrado, G.~S., and Dean, J.
\newblock Distributed representations of words and phrases and their
  compositionality.
\newblock \emph{Advances in neural information processing systems}, 2013.

\bibitem[Nathani et~al.(2019)Nathani, Chauhan, Sharma, and
  Kaul]{nathani2019learning}
Nathani, D., Chauhan, J., Sharma, C., and Kaul, M.
\newblock Learning attention-based embeddings for relation prediction in
  knowledge graphs.
\newblock \emph{arXiv preprint arXiv:1906.01195}, 2019.

\bibitem[Park et~al.(2019)Park, Kim, Lu, and Cho]{park2019relational}
Park, W., Kim, D., Lu, Y., and Cho, M.
\newblock Relational knowledge distillation.
\newblock In \emph{Proceedings of the IEEE/CVF Conference on Computer Vision
  and Pattern Recognition}, 2019.

\bibitem[Perozzi et~al.(2014)Perozzi, Al-Rfou, and Skiena]{perozzi2014deepwalk}
Perozzi, B., Al-Rfou, R., and Skiena, S.
\newblock Deepwalk: Online learning of social representations.
\newblock In \emph{Proceedings of the 20th ACM SIGKDD international conference
  on Knowledge discovery and data mining}, 2014.

\bibitem[Philip et~al.(2010)Philip, Han, and Faloutsos]{philip2010link}
Philip, S.~Y., Han, J., and Faloutsos, C.
\newblock \emph{Link mining: Models, algorithms, and applications}.
\newblock Springer, 2010.

\bibitem[Phuong \& Lampert(2019)Phuong and Lampert]{phuong2019towards}
Phuong, M. and Lampert, C.
\newblock Towards understanding knowledge distillation.
\newblock In \emph{International Conference on Machine Learning}. PMLR, 2019.

\bibitem[Reddi et~al.(2021)Reddi, Pasumarthi, Menon, Rawat, Yu, Kim, Veit, and
  Kumar]{reddi2021rankdistil}
Reddi, S., Pasumarthi, R.~K., Menon, A., Rawat, A.~S., Yu, F., Kim, S., Veit,
  A., and Kumar, S.
\newblock Rankdistil: Knowledge distillation for ranking.
\newblock In \emph{International Conference on Artificial Intelligence and
  Statistics}. PMLR, 2021.

\bibitem[Romero et~al.(2014)Romero, Ballas, Kahou, Chassang, Gatta, and
  Bengio]{romero2014fitnets}
Romero, A., Ballas, N., Kahou, S.~E., Chassang, A., Gatta, C., and Bengio, Y.
\newblock Fitnets: Hints for thin deep nets.
\newblock \emph{arXiv preprint arXiv:1412.6550}, 2014.

\bibitem[Sankar et~al.(2021)Sankar, Liu, Yu, and Shah]{sankar2021graph}
Sankar, A., Liu, Y., Yu, J., and Shah, N.
\newblock Graph neural networks for friend ranking in large-scale social
  platforms.
\newblock In \emph{Proceedings of the Web Conference}, pp.\  2535--2546, 2021.

\bibitem[Schlichtkrull et~al.(2018)Schlichtkrull, Kipf, Bloem, Berg, Titov, and
  Welling]{schlichtkrull2018modeling}
Schlichtkrull, M., Kipf, T.~N., Bloem, P., Berg, R. v.~d., Titov, I., and
  Welling, M.
\newblock Modeling relational data with graph convolutional networks.
\newblock In \emph{European semantic web conference}, pp.\  593--607. Springer,
  2018.

\bibitem[Shchur et~al.(2018)Shchur, Mumme, Bojchevski, and
  G{\"u}nnemann]{shchur2018pitfalls}
Shchur, O., Mumme, M., Bojchevski, A., and G{\"u}nnemann, S.
\newblock Pitfalls of graph neural network evaluation.
\newblock \emph{arXiv preprint arXiv:1811.05868}, 2018.

\bibitem[Shiao \& Papalexakis(2021)Shiao and
  Papalexakis]{shiao2021adversarially}
Shiao, W. and Papalexakis, E.~E.
\newblock Adversarially generating rank-constrained graphs.
\newblock In \emph{2021 IEEE 8th International Conference on Data Science and
  Advanced Analytics (DSAA)}. IEEE, 2021.

\bibitem[Shiao et~al.(2023)Shiao, Guo, Zhao, Papalexakis, Liu, and
  Shah]{shiao2022link}
Shiao, W., Guo, Z., Zhao, T., Papalexakis, E.~E., Liu, Y., and Shah, N.
\newblock Link prediction with non-contrastive learning.
\newblock In \emph{International Conference on Learning Representations}, 2023.

\bibitem[Tailor et~al.(2020)Tailor, Fernandez-Marques, and
  Lane]{tailor2020degree}
Tailor, S.~A., Fernandez-Marques, J., and Lane, N.~D.
\newblock Degree-quant: Quantization-aware training for graph neural networks.
\newblock \emph{arXiv preprint arXiv:2008.05000}, 2020.

\bibitem[Tang et~al.(2022)Tang, Liu, He, Wang, and Shah]{tang2022friend}
Tang, X., Liu, Y., He, X., Wang, S., and Shah, N.
\newblock Friend story ranking with edge-contextual local graph convolutions.
\newblock In \emph{Proceedings of the Fifteenth ACM International Conference on
  Web Search and Data Mining}, pp.\  1007--1015, 2022.

\bibitem[Tian et~al.(2022)Tian, Zhang, Guo, Zhang, and Chawla]{tian2022nosmog}
Tian, Y., Zhang, C., Guo, Z., Zhang, X., and Chawla, N.~V.
\newblock Nosmog: Learning noise-robust and structure-aware mlps on graphs.
\newblock \emph{arXiv preprint arXiv:2208.10010}, 2022.

\bibitem[Trouillon et~al.(2016)Trouillon, Welbl, Riedel, Gaussier, and
  Bouchard]{trouillon2016complex}
Trouillon, T., Welbl, J., Riedel, S., Gaussier, {\'E}., and Bouchard, G.
\newblock Complex embeddings for simple link prediction.
\newblock In \emph{International conference on machine learning}, pp.\
  2071--2080. PMLR, 2016.

\bibitem[Tsitsulin et~al.(2018)Tsitsulin, Mottin, Karras, and
  M{\"u}ller]{tsitsulin2018verse}
Tsitsulin, A., Mottin, D., Karras, P., and M{\"u}ller, E.
\newblock Verse: Versatile graph embeddings from similarity measures.
\newblock In \emph{Proceedings of the 2018 world wide web conference}, pp.\
  539--548, 2018.

\bibitem[Tung \& Mori(2019)Tung and Mori]{tung2019similarity}
Tung, F. and Mori, G.
\newblock Similarity-preserving knowledge distillation.
\newblock In \emph{Proceedings of the IEEE/CVF International Conference on
  Computer Vision}, 2019.

\bibitem[Vashishth et~al.(2020)Vashishth, Sanyal, Nitin, and
  Talukdar]{vashishth2020composition}
Vashishth, S., Sanyal, S., Nitin, V., and Talukdar, P.
\newblock Composition-based multi-relational graph convolutional networks.
\newblock \emph{arXiv preprint arXiv:1911.03082}, 2020.

\bibitem[Veli{\v{c}}kovi{\'c} et~al.(2017)Veli{\v{c}}kovi{\'c}, Cucurull,
  Casanova, Romero, Lio, and Bengio]{velivckovic2017graph}
Veli{\v{c}}kovi{\'c}, P., Cucurull, G., Casanova, A., Romero, A., Lio, P., and
  Bengio, Y.
\newblock Graph attention networks.
\newblock \emph{arXiv preprint arXiv:1710.10903}, 2017.

\bibitem[Wang et~al.(2020{\natexlab{a}})Wang, Shen, Huang, Wu, Dong, and
  Kanakia]{wang2020microsoft}
Wang, K., Shen, Z., Huang, C., Wu, C.-H., Dong, Y., and Kanakia, A.
\newblock Microsoft academic graph: When experts are not enough.
\newblock \emph{Quantitative Science Studies}, 2020{\natexlab{a}}.

\bibitem[Wang et~al.(2020{\natexlab{b}})Wang, Fu, Liao, Wang, Lei, and
  Mei]{wang2020exclusivity}
Wang, X., Fu, T., Liao, S., Wang, S., Lei, Z., and Mei, T.
\newblock Exclusivity-consistency regularized knowledge distillation for face
  recognition.
\newblock In \emph{European Conference on Computer Vision}, 2020{\natexlab{b}}.

\bibitem[Wang et~al.(2021)Wang, Zhou, Hong, Zou, and Su]{wang2021pairwise}
Wang, Z., Zhou, Y., Hong, L., Zou, Y., and Su, H.
\newblock Pairwise learning for neural link prediction.
\newblock \emph{arXiv preprint arXiv:2112.02936}, 2021.

\bibitem[Xu et~al.(2018)Xu, Li, Tian, Sonobe, Kawarabayashi, and
  Jegelka]{xu2018representation}
Xu, K., Li, C., Tian, Y., Sonobe, T., Kawarabayashi, K.-i., and Jegelka, S.
\newblock Representation learning on graphs with jumping knowledge networks.
\newblock In \emph{International conference on machine learning}, pp.\
  5453--5462. PMLR, 2018.

\bibitem[Yan et~al.(2020)Yan, Wang, Guo, and Lou]{yan2020tinygnn}
Yan, B., Wang, C., Guo, G., and Lou, Y.
\newblock Tinygnn: Learning efficient graph neural networks.
\newblock In \emph{Proceedings of the 26th ACM SIGKDD International Conference
  on Knowledge Discovery \& Data Mining}, 2020.

\bibitem[Yang et~al.(2021)Yang, Liu, and Shi]{yang2021extract}
Yang, C., Liu, J., and Shi, C.
\newblock Extract the knowledge of graph neural networks and go beyond it: An
  effective knowledge distillation framework.
\newblock In \emph{Proceedings of the Web Conference 2021}, 2021.

\bibitem[Yang et~al.(2020{\natexlab{a}})Yang, Qiu, Song, Tao, and
  Wang]{yang2020distilling}
Yang, Y., Qiu, J., Song, M., Tao, D., and Wang, X.
\newblock Distilling knowledge from graph convolutional networks.
\newblock In \emph{Proceedings of the IEEE/CVF Conference on Computer Vision
  and Pattern Recognition}, 2020{\natexlab{a}}.

\bibitem[Yang et~al.(2016)Yang, Cohen, and Salakhudinov]{yang2016revisiting}
Yang, Z., Cohen, W., and Salakhudinov, R.
\newblock Revisiting semi-supervised learning with graph embeddings.
\newblock In \emph{International conference on machine learning}, 2016.

\bibitem[Yang et~al.(2020{\natexlab{b}})Yang, Shou, Gong, Lin, and
  Jiang]{yang2020model}
Yang, Z., Shou, L., Gong, M., Lin, W., and Jiang, D.
\newblock Model compression with two-stage multi-teacher knowledge distillation
  for web question answering system.
\newblock In \emph{Proceedings of the 13th International Conference on Web
  Search and Data Mining}, 2020{\natexlab{b}}.

\bibitem[Yin et~al.(2022)Yin, Zhang, Wang, Wang, and Li]{yin2022algorithm}
Yin, H., Zhang, M., Wang, Y., Wang, J., and Li, P.
\newblock Algorithm and system co-design for efficient subgraph-based graph
  representation learning.
\newblock \emph{arXiv preprint arXiv:2202.13538}, 2022.

\bibitem[Ying et~al.(2018{\natexlab{a}})Ying, He, Chen, Eksombatchai, Hamilton,
  and Leskovec]{ying2018graph}
Ying, R., He, R., Chen, K., Eksombatchai, P., Hamilton, W.~L., and Leskovec, J.
\newblock Graph convolutional neural networks for web-scale recommender
  systems.
\newblock In \emph{Proceedings of the 24th ACM SIGKDD international conference
  on knowledge discovery \& data mining}, pp.\  974--983, 2018{\natexlab{a}}.

\bibitem[Ying et~al.(2018{\natexlab{b}})Ying, You, Morris, Ren, Hamilton, and
  Leskovec]{ying2018hierarchical}
Ying, Z., You, J., Morris, C., Ren, X., Hamilton, W., and Leskovec, J.
\newblock Hierarchical graph representation learning with differentiable
  pooling.
\newblock \emph{Advances in neural information processing systems},
  2018{\natexlab{b}}.

\bibitem[You et~al.(2018)You, Ying, Ren, Hamilton, and
  Leskovec]{you2018graphrnn}
You, J., Ying, R., Ren, X., Hamilton, W., and Leskovec, J.
\newblock Graphrnn: Generating realistic graphs with deep auto-regressive
  models.
\newblock In \emph{International conference on machine learning}. PMLR, 2018.

\bibitem[Yun et~al.(2021)Yun, Kim, Lee, Kang, and Kim]{yun2021neo}
Yun, S., Kim, S., Lee, J., Kang, J., and Kim, H.~J.
\newblock Neo-gnns: Neighborhood overlap-aware graph neural networks for link
  prediction.
\newblock \emph{Advances in Neural Information Processing Systems}, 2021.

\bibitem[Zeng et~al.(2019)Zeng, Zhou, Srivastava, Kannan, and
  Prasanna]{zeng2019graphsaint}
Zeng, H., Zhou, H., Srivastava, A., Kannan, R., and Prasanna, V.
\newblock Graphsaint: Graph sampling based inductive learning method.
\newblock \emph{arXiv preprint arXiv:1907.04931}, 2019.

\bibitem[Zhang et~al.(2020{\natexlab{a}})Zhang, Yao, Huang, Jiang, Li, and
  Chawla]{zhang2020few}
Zhang, C., Yao, H., Huang, C., Jiang, M., Li, Z., and Chawla, N.~V.
\newblock Few-shot knowledge graph completion.
\newblock In \emph{Proceedings of the AAAI Conference on Artificial
  Intelligence}, 2020{\natexlab{a}}.

\bibitem[Zhang et~al.(2020{\natexlab{b}})Zhang, Huang, Liu, Hu, Song, Ge,
  Zhang, Wang, Zhou, Shuang, et~al.]{zhang2020agl}
Zhang, D., Huang, X., Liu, Z., Hu, Z., Song, X., Ge, Z., Zhang, Z., Wang, L.,
  Zhou, J., Shuang, Y., et~al.
\newblock Agl: a scalable system for industrial-purpose graph machine learning.
\newblock \emph{arXiv preprint arXiv:2003.02454}, 2020{\natexlab{b}}.

\bibitem[Zhang et~al.(2020{\natexlab{c}})Zhang, Lin, Liu, Zhou, Tang, Liang,
  and Xing]{zhang2020iterative}
Zhang, H., Lin, S., Liu, W., Zhou, P., Tang, J., Liang, X., and Xing, E.~P.
\newblock Iterative graph self-distillation.
\newblock \emph{arXiv preprint arXiv:2010.12609}, 2020{\natexlab{c}}.

\bibitem[Zhang \& Chen(2017)Zhang and Chen]{zhang2017weisfeiler}
Zhang, M. and Chen, Y.
\newblock Weisfeiler-lehman neural machine for link prediction.
\newblock In \emph{Proceedings of the 23rd ACM SIGKDD international conference
  on knowledge discovery and data mining}, pp.\  575--583, 2017.

\bibitem[Zhang \& Chen(2018)Zhang and Chen]{zhang2018link}
Zhang, M. and Chen, Y.
\newblock Link prediction based on graph neural networks.
\newblock \emph{Advances in neural information processing systems}, 2018.

\bibitem[Zhang et~al.(2018)Zhang, Cui, Neumann, and Chen]{zhang2018end}
Zhang, M., Cui, Z., Neumann, M., and Chen, Y.
\newblock An end-to-end deep learning architecture for graph classification.
\newblock In \emph{Proceedings of the AAAI conference on artificial
  intelligence}, 2018.

\bibitem[Zhang et~al.(2021{\natexlab{a}})Zhang, Li, Xia, Wang, and
  Jin]{zhang2021labeling}
Zhang, M., Li, P., Xia, Y., Wang, K., and Jin, L.
\newblock Labeling trick: A theory of using graph neural networks for
  multi-node representation learning.
\newblock \emph{Advances in Neural Information Processing Systems},
  34:\penalty0 9061--9073, 2021{\natexlab{a}}.

\bibitem[Zhang et~al.(2021{\natexlab{b}})Zhang, Liu, Sun, and
  Shah]{zhang2021graph}
Zhang, S., Liu, Y., Sun, Y., and Shah, N.
\newblock Graph-less neural networks: Teaching old mlps new tricks via
  distillation.
\newblock \emph{arXiv preprint arXiv:2110.08727}, 2021{\natexlab{b}}.

\bibitem[Zhang et~al.(2020{\natexlab{d}})Zhang, Miao, Shao, Jiang, Chen, Ruas,
  and Cui]{zhang2020reliable}
Zhang, W., Miao, X., Shao, Y., Jiang, J., Chen, L., Ruas, O., and Cui, B.
\newblock Reliable data distillation on graph convolutional network.
\newblock In \emph{Proceedings of the 2020 ACM SIGMOD International Conference
  on Management of Data}, 2020{\natexlab{d}}.

\bibitem[Zhang et~al.(2020{\natexlab{e}})Zhang, Cui, and Zhu]{zhang2020deep}
Zhang, Z., Cui, P., and Zhu, W.
\newblock Deep learning on graphs: A survey.
\newblock \emph{IEEE Transactions on Knowledge and Data Engineering},
  2020{\natexlab{e}}.

\bibitem[Zhao et~al.(2021{\natexlab{a}})Zhao, Jin, Akoglu, and
  Shah]{zhao2021stars}
Zhao, L., Jin, W., Akoglu, L., and Shah, N.
\newblock From stars to subgraphs: Uplifting any gnn with local structure
  awareness.
\newblock \emph{arXiv preprint arXiv:2110.03753}, 2021{\natexlab{a}}.

\bibitem[Zhao et~al.(2021{\natexlab{b}})Zhao, Liu, Neves, Woodford, Jiang, and
  Shah]{zhao2021data}
Zhao, T., Liu, Y., Neves, L., Woodford, O., Jiang, M., and Shah, N.
\newblock Data augmentation for graph neural networks.
\newblock In \emph{Proceedings of the AAAI Conference on Artificial
  Intelligence}, 2021{\natexlab{b}}.

\bibitem[Zhao et~al.(2022{\natexlab{a}})Zhao, Jin, Liu, Wang, Liu, Günneman,
  Shah, and Jiang]{zhao2022graph}
Zhao, T., Jin, W., Liu, Y., Wang, Y., Liu, G., Günneman, S., Shah, N., and
  Jiang, M.
\newblock Graph data augmentation for graph machine learning: A survey.
\newblock \emph{arXiv preprint arXiv:2202.08871}, 2022{\natexlab{a}}.

\bibitem[Zhao et~al.(2022{\natexlab{b}})Zhao, Liu, Wang, Yu, and
  Jiang]{zhao2022learning}
Zhao, T., Liu, G., Wang, D., Yu, W., and Jiang, M.
\newblock Learning from counterfactual links for link prediction.
\newblock In \emph{International Conference on Machine Learning}, pp.\
  26911--26926. PMLR, 2022{\natexlab{b}}.

\bibitem[Zhao et~al.(2022{\natexlab{c}})Zhao, Tang, Zhang, Jiang, Rao, Song,
  Agrawal, Subbian, Yin, and Jiang]{zhaoautogda}
Zhao, T., Tang, X., Zhang, D., Jiang, H., Rao, N., Song, Y., Agrawal, P.,
  Subbian, K., Yin, B., and Jiang, M.
\newblock Autogda: Automated graph data augmentation for node classification.
\newblock In \emph{The First Learning on Graphs Conference},
  2022{\natexlab{c}}.

\bibitem[Zhao et~al.(2020)Zhao, Wang, Bates, Mullins, Jamnik, and
  Lio]{zhao2020learned}
Zhao, Y., Wang, D., Bates, D., Mullins, R., Jamnik, M., and Lio, P.
\newblock Learned low precision graph neural networks.
\newblock \emph{arXiv preprint arXiv:2009.09232}, 2020.

\bibitem[Zheng et~al.(2021)Zheng, Huang, Rao, Katariya, Wang, and
  Subbian]{zheng2021cold}
Zheng, W., Huang, E.~W., Rao, N., Katariya, S., Wang, Z., and Subbian, K.
\newblock Cold brew: Distilling graph node representations with incomplete or
  missing neighborhoods.
\newblock \emph{arXiv preprint arXiv:2111.04840}, 2021.

\bibitem[Zhou et~al.(2021)Zhou, Srivastava, Zeng, Kannan, and
  Prasanna]{zhou2021accelerating}
Zhou, H., Srivastava, A., Zeng, H., Kannan, R., and Prasanna, V.
\newblock Accelerating large scale real-time gnn inference using channel
  pruning.
\newblock \emph{arXiv preprint arXiv:2105.04528}, 2021.

\bibitem[Zhu et~al.(2021)Zhu, Zhang, Xhonneux, and Tang]{zhu2021neural}
Zhu, Z., Zhang, Z., Xhonneux, L.-P., and Tang, J.
\newblock Neural bellman-ford networks: A general graph neural network
  framework for link prediction.
\newblock \emph{Advances in Neural Information Processing Systems}, 2021.

\end{thebibliography}
\bibliographystyle{icml2023}

\clearpage
\appendix

\section{Further Related Work}

\begin{table*}[t]
\centering
\caption{Detailed statistics of datasets splits under production setting.}
\label{tab:data}
\vspace{-0.1in}
\begin{tabular}{l|cc|ccc} 
\toprule
& \multicolumn{2}{c|}{Nodes} & \multicolumn{3}{c}{Testing Edges} \\
        & \# Existing    & \# New    & \# Existing -- Existing & \# Existing -- New  & \# New -- New \\ 
\midrule
\cora    &1,896   &812   &765    &675    &142\\
\citeseer    &2,329   &998    &673    &568    &124\\
\pubmed  &15,774  &3,943   &5,648   &2,858   &358\\
\cs &14,666  &3,667   &10,482  &5,221   &675\\   
\physics    &27,594  &6,899   &31,399  &16,126  &2,067\\      
\computers    &11,002  &2,750   &31,095  &16,033  &2,043\\    
\photos   &6,120   &1,530   &15,248  &7,618   &950\\
\bottomrule
\end{tabular}
\end{table*}

\label{sec:related}


\textbf{Graph Neural Networks (GNNs).} Many GNN architectures have been proposed in recent years to model attributed graph data; most architectures follow the message passing \citep{gilmer2017neural,ying2018hierarchical,guo2021few, ma2021unified,liu2021elastic,liu2022graph,ju2023multi} paradigm. Different GNN customizations include degree normalization \citep{kipf2016semi}, neighbor sampling and neighbor separation \citep{hamilton2017inductive,zhao2021data}, self-attention \citep{velivckovic2017graph}, residual connections \citep{xu2018representation}, and more. \citet{alon2020bottleneck} proposed to use a fully-adjacent layer at the end of GNN to deal with the bottleneck problem of GNNs. Moreover, researchers also proposed subgraph-based methods \citep{bevilacqua2021equivariant, zhao2021stars}, tensor-based methods \citep{maron2019provably, geerts2022expressiveness}, and augmentation methods~\citep{zhao2022graph,liu2022graph,zhaoautogda} for improving GNNs.

\textbf{Link Prediction.} Link prediction has achieved great attention from the research community, considering its wide applications. Heuristic methods~\citep{philip2010link} were proposed to make the link prediction by measuring the link scores based on the structure information, such as the common neighbors and the shortest path. 2-order~\citep{adamic2003friends} and high-order~\citep{brin2012reprint, jeh2002simrank} heuristic methods were proposed to further improve the effectiveness. In recent years, GNN-based methods~\citep{zhang2018link,yun2021neo,zhao2022learning} showed their promising performances for link prediction.
One line of work follows the node embedding-based strategy, as previously discussed in \cref{sec: prelimi}, where the GNN-based encoder learns node representations and the decoder predicts whether the link exists. 
It is worth mentioning that knowledge graph completion follows this strategy to predict not only the link existence but also the type of the link~\citep{schlichtkrull2018modeling,nathani2019learning,vashishth2020composition,zhang2020few}. The knowledge graph completion methods mainly use heterogeneous graph neural networks, which are sensitive to different edge types.

Another line of work casts link prediction tasks to binary classification problems on the enclosing subgraphs around each node pair~\citep{zhang2018link, cai2020multi,cai2021line}. Although these methods can improve task performance, they are usually computationally expensive and cannot scale well in practical use-cases \citep{yin2022algorithm}. Similarly, \citet{zhu2021neural} proposed a GNN link prediction paradigm by encoding information of all paths between two nodes, which is also very expensive. 

\textbf{GNN Inference Acceleration.} Pruning~\citep{zhou2021accelerating, chen2021unified} and quantization~\citep{zhao2020learned, tailor2020degree} strategies were proposed for accelerating GNN inference. These methods do accelerate GNNs, but they rely on graph data for message passing and thus leave much room for speed improvement. We note that these approaches are complementary to cross-model distillation, and can be employed together with KD for additional inference time improvements. Other than the above acceleration methods, \citet{hu2021graph} and \citet{zhang2021graph} accelerated GNNs by distilling them to MLP. These works focus on KD for node classification tasks, whereas we focus on link prediction tasks. GNNAutoScale~\citep{fey2021gnnautoscale} proposed an effective method to accelerate the training process of GNNs. It also reduces the inference to a constant factor by directly using historical embeddings stored offline. However, in this case, all the methods in~\cref{fig:inference} can share the same inference time benefits. Moreover, GNNAutoScale is not suitable for the production setting, where new nodes (without historical embeddings) appear frequently after the training process. So we did not include it as a baseline in this work.

\textbf{Knowledge Distillation (KD).} Logit-based~\citep{hinton2015distilling, furlanello2018born, zhang2021graph} and representation-based~\citep{romero2014fitnets, gou2021knowledge} matching are two common KD methods, which match final-layer and intermediate-layer predicted logits between the teacher and the student, respectively. Our work is the first to adapt and evaluate these approaches in the link prediction setting, to the best of our knowledge.

For representation-based KD, several work~\citep{park2019relational, tung2019similarity, joshi2021representation} proposed relational KD, which corresponds to instance-to-instance KD while preserving metrics among representations of similar instances. For GNNs, \cite{yang2020distilling} used knowledge of the neighboring nodes to teach the student to better classify the center node. In contrast, our KD strategies focus on transferring relational knowledge between each pair of nodes from teacher to student. Both the rank-matching and distribution-matching strategies help the student to better capture the relational graph topology information and make better link predictions. 

RankDistill~\citep{reddi2021rankdistil} and Topology Distillation~\citep{kang2021topology} are designed to transfer ranking knowledge from the teacher to the student. Different from our work which distill the relational information in a graph context, they distill ranking in a non-graph context between teacher and student. We adopt different sampling and matching methods based on our different motivations. Further analysis is shown in~\cref{sec:apdx-distillation-baselines}.

\textbf{Knowledge Distillation on GNNs.} Existing GNN-based KD work are mostly based on the logit-based KD~\citep{hinton2015distilling} to obtain light-weight models~\citep{zhang2020reliable, zheng2021cold, yang2021extract}. \citet{yan2020tinygnn} proposed to train a student GNN with fewer parameters using KD. \citet{yang2021extract} improved the designed student model, which consists of label propagation and feature-based prior knowledge, using the pre-trained teacher GNN. Different from the above work, LSP~\citep{yang2020distilling} and G-CRD~\citep{joshi2021representation} proposed structure-preserving KD methods, which are specifically designed for GNN. Both of these work follow the original relational KD to preserve the metrics among node representations and are applied on node classification tasks.

\section{Additional Datasets Details}
\label{sec:apdx-data}
Here we present the details of the datasets used in the experiments. \cora, \citeseer, and \pubmed~\citep{yang2016revisiting} are all representative citation network datasets, where the nodes and edges represent papers and citations, respectively. \cs, \physics~\citep{shchur2018pitfalls} and \collab~\citep{wang2020microsoft} are all collaboration networks based on MAG, where the nodes represent authors and the edges indicate the collaboration for the paper. \computers and \photos~\citep{shchur2018pitfalls} are two well-known co-purchased graphs~\citep{mcauley2015image}, where the nodes represent goods and the edges indicate two items were bought together. 

\section{Additional Evaluation Setting Details}
\label{sec:apdx-eval}
\subsection{Transductive Setting}
The transductive setting is a standard setting for link prediction~\citep{kipf2016variational,zhang2017weisfeiler,zhang2018link,yun2021neo,zhao2022learning}, where the nodes in training/validation/testing are all visible in the training graph, but subsets of positive links are masked out for validation and test sets. 

\begin{table*}[t]
\small
\centering
\caption{Detailed link prediction performance measured by Hits@20 under \textbf{production} setting. Best and second best performances are marked with bold and underline, respectively.}
\vspace{-0.1in}
\label{tab:product_hits_full}
\begin{tabular}{l|cc|cc|cccc}
\toprule
          & GNN     & MLP      & \kdp     & \kdf   & \ours &$\Delta_{DirectKD}$  & $\Delta_{MLP}$    & $\Delta_{GNN}$ \\
\midrule
\multicolumn{9}{c}{Overall}\\ 
\midrule                                    
\cora      & \ms{\underline{27.80}} {2.11}   & \ms{22.90} {2.22}  & \ms{22.65} {2.51} & \ms{22.24} {0.55} & \ms{\textbf{27.87}} {1.24} & 5.22  & 4.97  & 0.07   \\
\citeseer  & \ms{\textbf{38.78}} {2.59} & \ms{31.21} {3.75}  & \ms{29.35} {2.55} & \ms{26.23} {1.08} & \ms{\underline{34.75}} {2.45} & 5.40 & 3.54  & -4.03    \\
\pubmed    & \ms{\underline{52.71}} {1.81} & \ms{38.01} {1.67}  & \ms{39.03} {4.21} & \ms{43.27} {3.12} & \ms{\textbf{53.48}} {1.52} & 10.21 & 15.47 & 0.77    \\
\cs        & \ms{\underline{60.69}} {3.17} & \ms{38.15} {10.78} & \ms{48.07} {2.39} & \ms{58.90} {1.32} & \ms{\textbf{60.74}} {1.41} & 1.84 & 22.59 & 0.05    \\
\physics   & \ms{\textbf{55.82}} {2.43} & \ms{29.99} {1.96}  & \ms{22.74} {1.03} & \ms{36.32} {2.29} & \ms{\underline{52.83}} {1.50}  & 16.51 & 22.84 & -2.99   \\
\computers & \ms{\textbf{34.38}} {1.41} & \ms{19.43} {0.82}  & \ms{12.79} {1.43} & \ms{20.28} {1.01} & \ms{\underline{24.58}} {3.33} & 4.30 & 5.15  & -9.80      \\
\photos    & \ms{\textbf{51.03}} {6.05} & \ms{34.29} {2.49}  & \ms{24.63} {2.20}  & \ms{40.58} {1.63} & \ms{\underline{43.79}} {1.27} & 3.21 & 9.50  & -7.24   \\ 
\midrule

\multicolumn{9}{c}{Existing -- Existing }\\ 
\midrule                                    
\cora      & \ms{\underline{28.81}} {2.01} & \ms{28.00} {2.70}   & \ms{27.66} {3.01} & \ms{27.03} {0.65} & \ms{\textbf{33.31}} {1.29} & 5.65 & 5.31  & 4.5     \\
\citeseer  & \ms{\textbf{38.10}} {2.70}  & \ms{33.88} {3.50}   & \ms{32.24} {2.89} & \ms{27.52} {0.94} & \ms{\underline{37.50}} {2.43} & 5.26 & 3.62  & -0.60    \\
\pubmed    & \ms{\underline{52.67}}  {1.78} & \ms{41.58} {1.61}  & \ms{42.57} {4.32} & \ms{46.32} {3.08} & \ms{\textbf{57.16}} {1.34} & 10.84 & 15.58 & 4.49    \\
\cs        & \ms{\underline{61.52}} {3.10}  & \ms{40.27} {11.69} & \ms{50.78} {2.50}  & \ms{62.17} {1.45} & \ms{\textbf{63.99}} {1.36} & 1.82  & 23.72 & 2.47    \\
\physics   & \ms{\textbf{56.56}} {2.42} & \ms{32.32} {2.32}  & \ms{23.88} {1.14} & \ms{38.74} {2.50}  & \ms{\underline{56.04}} {1.47} & 17.30  & 23.72 & -0.52   \\
\computers & \ms{\textbf{35.13}} {1.48} & \ms{21.46} {1.08}  & \ms{13.81} {1.56} & \ms{22.78} {1.17} & \ms{\underline{26.89}} {3.60}  & 4.11 & 5.43  & -8.24    \\
\photos    & \ms{\textbf{51.90}} {6.24} & \ms{37.47} {2.73}  & \ms{26.54} {2.55} & \ms{44.51} {2.10}  & \ms{\underline{48.38}} {1.30}  & 3.87 & 10.91 & -3.52   \\  
\midrule

\multicolumn{9}{c}{Existing -- New}\\ 
\midrule                                    
\cora      & \ms{\textbf{25.78}} {2.33} & \ms{19.47} {2.09}  & \ms{19.11} {2.03} & \ms{18.58} {1.28} & \ms{\underline{23.08}} {1.51} & 3.97 & 3.61  & -2.7    \\
\citeseer  & \ms{\textbf{38.73}} {2.37} & \ms{30.77} {4.07}  & \ms{28.77} {2.70}  & \ms{26.65} {1.52} & \ms{\underline{34.30}} {2.40}  & 5.53 & 3.53  & -4.43   \\
\pubmed    & \ms{\textbf{53.98}} {2.29} & \ms{23.70} {2.09}  & \ms{24.91} {4.00} & \ms{32.21} {3.38} & \ms{\underline{38.94}} {2.44} & 6.73 & 15.24 & -15.04  \\
\cs        & \ms{\textbf{56.78}} {3.57} & \ms{29.25} {7.05}  & \ms{36.60} {2.17} & \ms{45.28} {0.93} & \ms{\underline{47.05}} {1.72} & 1.77 & 17.80  & -9.73    \\
\physics   & \ms{\textbf{52.90}} {2.44} & \ms{20.61} {1.01}  & \ms{18.23} {0.74} & \ms{26.57} {1.81} & \ms{\underline{39.73}} {1.75} & 13.16 & 19.12 & -13.17 \\
\computers & \ms{\textbf{31.07}} {1.17} & \ms{11.00} {1.37}  & \ms{8.53} {1.16} & \ms{9.85} {0.54} & \ms{\underline{14.88}} {2.58} & 5.03 & 3.88  & -16.19   \\
\photos    & \ms{\textbf{47.42}} {5.18} & \ms{21.00} {1.65}  & \ms{16.75} {0.92} & \ms{24.10} {1.38} & \ms{\underline{24.27}} {2.07} & 0.17 & 3.27  & -23.15   \\ 
\midrule

\multicolumn{9}{c}{New -- New}\\ 
\midrule                                    
\cora      & \ms{\textbf{31.97}} {6.65} & \ms{11.69} {2.19}  & \ms{12.54} {2.83} & \ms{13.80} {1.37} & \ms{\underline{16.90}} {5.50}  & 3.10 & 5.21  & -15.07   \\
\citeseer  & \ms{\textbf{42.74}} {4.49} & \ms{18.71} {4.54}  & \ms{16.29} {3.80}  & \ms{17.26} {3.54} & \ms{\underline{21.94}} {4.39} & 4.68 & 3.23  & -20.8   \\
\pubmed    & \ms{\textbf{33.18}} {1.24} & \ms{5.45}  {1.24}  & \ms{4.55}  {4.55} & \ms{11.36} {4.82} & 
\ms{\underline{15.00}} {6.35} & 3.64 & 9.55  & -18.18  \\
\cs        & \ms{\textbf{64.10}} {3.55} & \ms{26.27} {8.79}  & \ms{33.73} {3.81} & \ms{38.07} {2.90}  & \ms{\underline{42.89}} {1.83} & 4.82 & 16.62 & -21.21   \\
\physics   & \ms{\textbf{48.96}} {3.53} & \ms{13.20}  {1.62}  & \ms{12.56} {1.85} & \ms{18.88} {2.22} & \ms{\underline{32.80}} {1.55} & 13.92 & 19.6  & -16.16  \\
\computers & \ms{\textbf{32.61}} {1.89} & \ms{5.55} {1.56}  & \ms{6.72} {0.66} & \ms{3.87}  {1.24} & \ms{\underline{10.25}} {1.41} & 3.53 & 4.7   & -22.36   \\
\photos    & \ms{\textbf{43.54}} {6.6}  & \ms{10.09} {3.99}  & \ms{8.14}  {1.15} & \ms{12.21} {1.31} & \ms{\underline{14.87}} {3.09} & 2.66 & 4.78  & -28.67  \\
\bottomrule
\end{tabular}
\vspace{-0.1in}
\end{table*}

\subsection{Production Setting}
In this work, we design a new production setting to resemble the real-world link prediction scenario. This setting mimics practical link prediction use cases. For example, user friend recommendations on social platforms where new users (nodes) and friendships (links) appear frequently. Under the production setting, the newly occurred nodes and edges that can not be seen during the training stage would appear in the graph at inference time. 

Specifically, the following are the detailed procedures of splitting the datasets into the production setting:
\begin{itemize}
    \item \textbf{Split all nodes:} Given the graph $G=(\mathcal{V}, \mathcal{E})$, we randomly sample 10\% of nodes from $\mathcal{V}$ as the new nodes $\mathcal{V}^N$ and remove them from the training graph. We denote the remaining nodes by $\mathcal{V}^E$, where superscripts $E$ stands for \emph{Existing} and $N$ stands for \emph{New}. Note that for \cora and \citeseer, we sample 30\% nodes as new nodes because these two datasets are too small. 
    \item \textbf{Split all edges:} We then split the edges $\mathcal{E}$ according to the node splits into three sets: $\mathcal{E}^{E-E}$, $\mathcal{E}^{E-N}$, and $\mathcal{E}^{N-N}$, denoting the links between existing--existing, existing--new, and new--new node pairs, respectively.
    \item \textbf{Split edges in $\mathcal{E}^{E-E}$:} For the existing--existing node pairs, we split it into three sets following an 80/10/10 splitting ratio: 80\% as training edges, 10\% as new visible edges for message passing, and 10\% as testing edges. Note that validation set contains only existing nodes $\mathcal{V}^E$ as the new nodes are not visible during training. 
    \item \textbf{Split edges in $\mathcal{E}^{E-N}$ and $\mathcal{E}^{N-N}$:} We follow the same ratio and split these two sets following with 90/10 splitting ratio: 90\% as newly visible edges (used only for message passing during testing inference), and 10\% as testing edges. 
    \item \textbf{Message passing edges during training:} During training, the GNN model can only utilize the 80\% existing-existing training edges for message passing.
    \item \textbf{Message passing edges for inferencing:} During inference, the GNN model can conduct message passing on all edges except the testing ones. Specifically, the training and testing (total of 90\%) sets of $\mathcal{E}^{E-E}$, and the 90\% of newly visible message passing edges in $\mathcal{E}^{E-N}$ and $\mathcal{E}^{N-N}$.
    \item \textbf{Testing edges:} We test all methods on the above-mentioned three separate testing edge sets (10\% of each) sampled from $\mathcal{E}^{E-E}$, $\mathcal{E}^{E-N}$, and $\mathcal{E}^{N-N}$, respectively.
\end{itemize}
\cref{tab:data} shows the detailed statistics of different datasets under this setting.

\subsection{Cold Start Setting}
Followed by the production setting, we remove all the new edges appearing newly in the inference time. Then the new nodes will be the strict cold start nodes with no neighbor information for the model to predict. The experimental results shown in~\cref{sec:cold} are conducted with this setting.

\section{Additional Experimental Results}
\label{sec:apdx-results}

\begin{table*}[t]
\small
\centering
\caption{Link prediction performance measured by AUC under \textbf{transductive} setting.}
\label{tab:trans_auc}
\vspace{-0.1in}
\begin{tabular}{l|cc|cc|cccc} \toprule
          & GNN  & MLP   & \kdp   & \kdf   & \ours  & $\Delta_{DirectKD}$ & $\Delta_{MLP}$    & $\Delta_{GNN}$  \\ \midrule
\cora      & \ms{\underline{95.03}} {0.37} & \ms{94.80}  {0.44} & \ms{94.67} {0.58} & \ms{94.05} {0.17} & \ms{\textbf{95.23}} {0.49} & 0.56 & 0.43 & 0.20   \\ 
\citeseer  & \ms{\underline{95.15}} {0.58} & \ms{93.11} {1.21} & \ms{94.11} {0.21} & \ms{92.88} {0.37} & \ms{\textbf{95.32}} {0.21} & 1.21 & 2.21 & 0.17  \\ 
\pubmed    & \ms{93.84} {0.31} & \ms{97.89} {0.07} & \ms{97.82} {0.06} & \ms{\textbf{97.96}} {0.02} & \ms{\underline{97.90}} {0.09}  & -0.06 & 0.01 & 4.06\\ 
\cs        & \ms{97.43} {0.23} & \ms{97.61} {0.52} & \ms{98.05} {0.14} & \ms{\textbf{98.33}} {0.05} & \ms{\underline{98.06}} {0.04} & -0.27 & 0.45 & 0.63 \\ 
\physics   & \ms{98.80} {0.02} & \ms{98.71} {0.05} & \ms{98.36} {0.07} & \ms{\underline{98.96}} {0.02} & \ms{\textbf{99.10}}  {0.02} & 0.14 & 0.39 & 0.30   \\ 
\computers & \ms{\underline{98.76}} {0.03} & \ms{98.46} {0.08} & \ms{98.11} {0.14} & \ms{98.66} {0.06} & \ms{\textbf{98.84}} {0.09} & 0.18 & 0.38 & 0.08  \\ 
\photos    & \ms{\underline{98.98}} {0.02} & \ms{98.71} {0.08} & \ms{98.51} {0.06} & \ms{98.95} {0.04} & \ms{\textbf{99.03}} {0.06} & 0.08 & 0.32 & 0.05  \\
\bottomrule
\end{tabular}
\end{table*}

\begin{table*}[!t]
\small
\centering
\caption{Link prediction performance measured by AUC under \textbf{production} setting.}
\label{tab:product_auc}
\vspace{-0.1in}
\begin{tabular}{l|cc|cc|cccc}
\toprule
          & GNN  & MLP   & \kdp   & \kdf   & \ours  & $\Delta_{DirectKD}$ & $\Delta_{MLP}$    & $\Delta_{GNN}$  \\\midrule
\multicolumn{9}{c}{Overall}\\ \midrule                       
\cora      & \ms{72.59} {1.63} & \ms{\underline{73.41}} {2.04} & \ms{70.67} {1.62} & \ms{64.62} {0.51} & \ms{\textbf{78.22}} {1.14}  & 7.55 & 4.81  & 5.63  \\
\citeseer  & \ms{69.15} {1.82} & \ms{\underline{77.36}} {3.38} & \ms{75.04} {3.20}  & \ms{67.67} {0.59} & \ms{\textbf{80.13}} {0.98} & 5.09 & 2.77  & 10.98   \\
\pubmed    & \ms{90.45} {0.45} & \ms{96.07} {0.13} & \ms{\underline{96.13}} {0.26} & \ms{\textbf{96.74}} {0.05} & \ms{94.30}  {0.34}  & -2.44 & -1.77 & 3.85 \\
\cs        & \ms{\textbf{97.08}} {0.16} & \ms{95.96} {1.19} & \ms{96.59} {0.08} & \ms{96.76} {0.03} & \ms{\underline{96.87}} {0.03} & 0.11 & 0.91  & -0.21   \\
\physics   & \ms{\underline{98.60}} {0.02} & \ms{97.70} {0.04} & \ms{97.46} {0.08} & \ms{98.00} {0.01} & \ms{\textbf{98.75}} {0.11} & 0.75  & 1.05  & 0.15   \\
\computers & \ms{\textbf{98.67}} {0.05} & \ms{97.85} {0.04} & \ms{97.59} {0.07} & \ms{\underline{97.95}} {0.03} & \ms{97.89} {0.04} & -0.06 & 0.04  & -0.78 \\
\photos    & \ms{\textbf{98.78}} {0.14} & \ms{97.97} {0.08} & \ms{97.85} {0.06} & \ms{\underline{98.18}} {0.04} & \ms{98.05} {0.03} & -0.13 & 0.08  & -0.73  \\ \midrule
\multicolumn{9}{c}{Existing -- Existing}\\ \midrule      
\cora      & \ms{70.80} {2.14} & \ms{\underline{74.42}} {2.70}  & \ms{70.69} {2.00} & \ms{64.82} {0.75} & \ms{\textbf{78.43}} {1.44} & 7.74  & 4.01  & 7.63   \\
\citeseer  & \ms{67.34} {1.81} & \ms{\underline{76.83}} {3.41} & \ms{73.79} {3.12} & \ms{68.00} {2.03} & \ms{\textbf{78.36}} {1.41} & 4.57  & 1.53  & 11.02  \\
\pubmed    & \ms{90.44} {0.46} & \ms{96.69} {0.13} & \ms{\underline{96.72}} {0.21} & \ms{\textbf{97.24}} {0.05} & \ms{95.17} {0.33} & -2.07  & -1.52 & 4.73  \\
\cs        & \ms{\textbf{97.01}} {0.16} & \ms{96.08} {1.11} & \ms{96.70} {0.08} & \ms{96.91} {0.03} & \ms{\underline{97.00}} {0.03} & 0.09  & 0.92  & -0.01  \\
\physics   & \ms{\underline{98.60}} {0.02} & \ms{97.96} {0.05} & \ms{97.65} {0.09} & \ms{98.20} {0.02} & \ms{\textbf{98.76}} {0.16} & 0.56  & 0.80   & 0.16   \\
\computers & \ms{\textbf{98.70}} {0.05} & \ms{98.27} {0.05} & \ms{97.95} {0.09} & \ms{98.41} {0.03} & \ms{\underline{98.51}} {0.04} & 0.10  & 0.24  & -0.19   \\
\photos    & \ms{\textbf{98.80}} {0.14} & \ms{98.33} {0.09} & \ms{98.20} {0.07} & \ms{98.57} {0.07} & \ms{\underline{98.61}} {0.04} & 0.04 & 0.28  & -0.19  \\\midrule
\multicolumn{9}{c}{Existing -- New}\\ \midrule      
\cora      & \ms{\underline{72.61}} {1.50}  & \ms{72.06} {1.55} & \ms{70.18} {1.41} & \ms{64.07} {0.58} & \ms{\textbf{77.65}} {1.12} & 7.47 & 5.59  & 5.04    \\
\citeseer  & \ms{69.90} {1.88} & \ms{\underline{77.58}} {3.48} & \ms{76.05} {3.51} & \ms{67.13} {1.74} & \ms{\textbf{81.23}} {0.71} & 5.18 & 3.65  & 11.33  \\
\pubmed    & \ms{90.82} {0.38} & \ms{93.67} {0.23} & \ms{\underline{93.82}} {0.54} & \ms{\textbf{94.83}} {0.14} & \ms{90.97} {0.74} & -3.86 & -2.70  & 0.15  \\
\cs        & \ms{\textbf{97.31}} {0.20}  & \ms{95.46} {1.53} & \ms{96.18} {0.12} & \ms{96.18} {0.08} & \ms{\underline{96.31}} {0.10}  & 0.13 & 0.85  & -1.00     \\
\physics   & \ms{\textbf{98.57}} {0.04} & \ms{96.64} {0.08} & \ms{96.66} {0.09} & \ms{\underline{97.17}} {0.04} & \ms{95.72} {0.27} & -1.45 & -0.92 & -2.85  \\
\computers & \ms{\textbf{98.60}} {0.05} & \ms{\underline{96.23}} {0.07} & \ms{96.22} {0.07} & \ms{96.19} {0.08} & \ms{95.42} {0.08} & -0.80 & -0.81 & -3.18  \\
\photos    & \ms{\textbf{98.69}} {0.14} & \ms{96.53} {0.03} & \ms{96.45} {0.09} & \ms{\underline{96.60}} {0.08} & \ms{95.76} {0.16}  & -0.84 & -0.77 & -2.93\\\midrule
\multicolumn{9}{c}{New -- New}\\ \midrule      
\cora      & \ms{\textbf{82.10}} {1.57} & \ms{74.46} {1.40}  & \ms{72.85} {1.92} & \ms{66.12} {1.39} & \ms{\underline{79.85}} {1.30}  & 7.00 & 5.39  & -2.25      \\
\citeseer  & \ms{75.48} {1.67} & \ms{\underline{79.23}} {3.08} & \ms{77.13} {3.24} & \ms{68.36} {2.67} & \ms{\textbf{84.68}} {0.89}  & 7.55 & 5.45  & 9.20   \\
\pubmed    & \ms{84.30} {0.90}  & \ms{87.97} {1.02} & \ms{\underline{88.72}} {1.19} & \ms{\textbf{89.95}} {0.50}  & \ms{83.54} {2.57} & -6.41 & -4.43 & -0.76 \\
\cs        & \ms{\textbf{97.99}} {0.23} & \ms{95.03} {1.34} & \ms{95.39} {0.44} & \ms{95.22} {0.22} & \ms{\underline{95.97}} {0.34} & 0.58 & 0.94  & -2.02   \\
\physics   & \ms{\textbf{98.84}} {0.12} & \ms{95.72} {0.30}  & \ms{96.43} {0.20}  & \ms{\underline{96.80}} {0.24} & \ms{94.78} {0.34} & -2.02 & -0.94 & -4.06  \\
\computers & \ms{\textbf{98.07}} {0.10}  & \ms{93.22} {0.16} & \ms{\underline{93.30}} {0.32} & \ms{92.96} {0.39} & \ms{92.09} {0.73} & -1.21 & -1.13 & -5.98  \\
\photos    & \ms{\textbf{98.35}} {0.16} & \ms{94.21} {0.28} & \ms{\underline{94.69}} {0.09} & \ms{93.79} {0.43} & \ms{92.09} {0.59} & -2.60 & -2.12 & -6.26 \\
\bottomrule
\end{tabular}
\end{table*}

\subsection{Detailed Hits@20 Results under Production Setting}
\label{apdx:product_full}
Here, we stratify each method's performance on the three different categories of the test edges in \cref{tab:product_hits_full}. From these more stratified test performances, we observe that \ours can generally achieve similar performance with GNN on the existing--existing category, but much worse on the other two categories that involve newly appeared nodes. We hypothesize that this is because GNN neighbor aggregation improves generalization for low-degree nodes. We also observe that the performance gaps between the teacher GNN and stand-alone MLP on new--new and existing--new are much larger than that of the existing--existing category, which also evidences that GNNs have better inductive bias than MLPs on graph data. Nonetheless, we note that such a significant and consistent performance improvement of \ours over MLP is valuable for large-scale industrial applications, given their popularity in practice. 

Nonetheless, while obtaining accurate link prediction for new nodes (e.g. recommending friends to new users) is very important in the production scenario, it is equally and perhaps even more important to achieve stronger prediction performance on the existing nodes (e.g. recommending friends to existing users in the real-world link prediction scenario). This is because, in practice, we have many large graphs (e.g. social graphs) which grow slowly, but which require frequent predictions for existing nodes (e.g. still have a large and active user-base to serve) -- for example, Facebook's user growth\footnote{\url{https://www.oberlo.com/statistics/how-many-users-does-facebook-have}} reveals that the platform added 51 million new users in 2022, but still had a total of 2.96 billion users that year. Additionally, Twitter experienced a decline in its user base in 2022\footnote{\url{https://www.statista.com/statistics/238729/twitters-annual-growth-rate-in-the-us/}}. Given these circumstances, “Overall” and “Existing-Existing” still matter a lot in practical settings because they account for this large cohort.

\subsection{Link Prediction Results Measured by AUC under Transductive and Production Settings}
We present AUC results on all the non-OGB datasets under the transductive setting in~\cref{tab:trans_auc} and production setting in~\cref{tab:product_auc}. In~\cref{tab:trans_auc}, we observe that our method outperforms both MLP and the teacher GNN on all the datasets under the transductive setting. For the production setting (\cref{tab:product_auc}), our method achieves better performances than the teacher GNN on 4/7 datasets.

\begin{figure}[t]
    \centering
    \includegraphics[ width=0.48\textwidth]{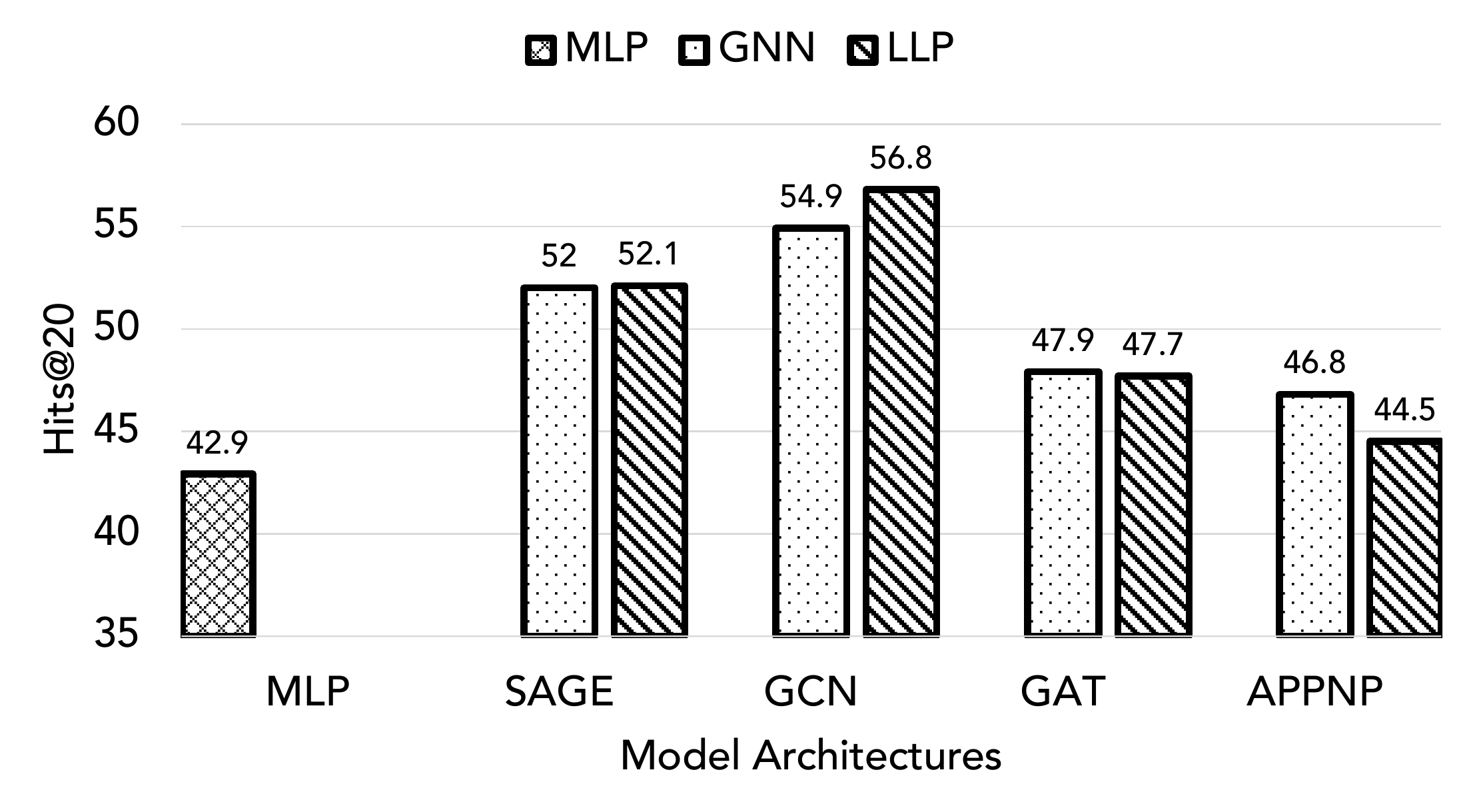}
    \vspace{-0.1in}
    \caption{Link prediction performance measured by Hits@20 on \pubmed under transductive setting with different GNN teachers (SAGE, GAT, GCN, and APPNP).}
    \label{fig:teacher}
\end{figure}

\subsection{Performance with Different Teachers}
\label{apdx:diff_teacher}
In our experiments, we use SAGE as the teacher GNN under both transductive and production settings. We further evaluate \ours's performance with other GNNs as teachers, such as GCN, GAT, and APPNP. In \cref{fig:teacher}, we observe that our method consistently outperforms MLP and the corresponding teacher GNNs. However, absolute performance is closely related to the teacher GNNs.

\begin{table}[t]
\centering
\caption{Link prediction performance measured by Hits@20 of SAGE, PLNLP~\citep{wang2021pairwise} and \ours. \ours(SAGE) indicates \ours with SAGE as the teacher. \ours(PLNLP) indicates \ours with PLNLP as the teacher.}
\label{tab:plnlp_hits}
\vspace{-0.1in}
\scalebox{0.83}{
\begin{tabular}{l|cc|cc}
\toprule
          & SAGE  & \ours(SAGE) & PLNLP & \ours(PLNLP) \\\midrule
\cs        & 59.51 & 68.62     & 63.42 & 69.76      \\\midrule
\physics   & 66.74 & 72.01     & 67.18 & 71.71      \\\midrule
\computers & 31.66 & 35.32     & 32.92 & 35.55      \\\midrule
\photos    & 51.50  & 49.32     & 52.01 & 51.97      \\\midrule
\collab    & 48.69 & 49.10      & 54.60  & 51.14\\
\bottomrule
\end{tabular}}
\end{table} 

\subsection{Comparison among SAGE, PLNLP, and \ours}\label{apdx:plnlp}
\textbf{Effectiveness Comparison.} Here we extend the experiments in \cref{fig:teacher} with more datasets and also a more advanced teacher model PLNLP~\citep{wang2021pairwise}, which is ranked \#6 on the official \collab leaderboard. \cref{tab:plnlp_hits} shows the performance of different GNNs and \ours with different GNNs as teacher models. We can observe that \ours with PLNLP as teacher model achieves higher link prediction performances than the \ours with SAGE as teacher model, showing \ours is able to learn from more advanced teacher GNNs. 

\begin{table}[t]
\centering
\caption{Number of model parameters, inference time, and prediction performance comparison of SAGE, PLNLP~\citep{wang2021pairwise} and \ours on \collab.}
\label{tab:plnlp_params}
\vspace{-0.1in}
\scalebox{0.75}{
\begin{tabular}{l|cc|cc}
\toprule
                   & SAGE   & \ours(SAGE) & PLNLP    & \ours(PLNLP) \\\midrule
\# Params      & 263,169 & 2,232,321   & 60,644,864 & 2,232,321    \\\midrule
Inference Time & 156.3  & 8.1       & 143.2    & 8.1        \\\midrule
Hits@50        & 48.69  & 49.1      & 54.6     & 51.14 \\
\bottomrule
\end{tabular}}
\end{table}

\textbf{Efficiency Comparison.} Due to the dependency on the graph structure, the inference time required for GNN methods can be very large when compared with that of MLPs. To illustrate, we show empirical results on \collab, in which we compare SAGE, PLNLP and \ours in terms of their number of model parameters, inference time, and prediction performance (Hits@50). In \cref{tab:plnlp_params}, we found that the two GNNs have high delay during inference (over 100-millisecond for a single node). In many real-time applications, such high latency would make the deployment of GNNs infeasible given business or engineering constraints, considering that many such queries may occur simultaneously in industrial settings. On the other hand, we observe that \ours can preserve most of the performance with significantly less time, indicating \ours is more suitable for certain contexts in which time constraints are important.

\begin{figure*}[t]
    \centering
    {\includegraphics[ width=0.48\textwidth]{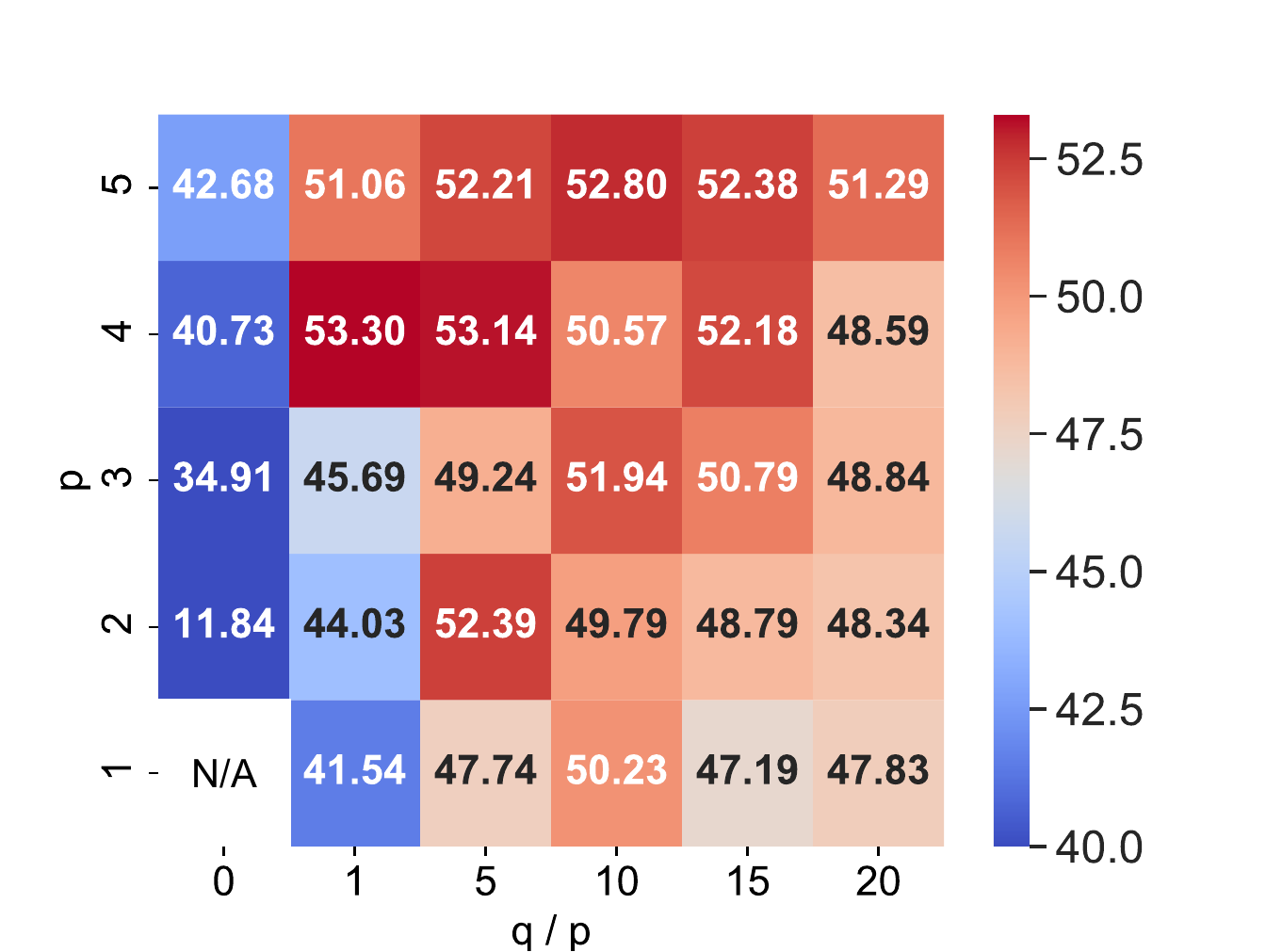}\label{fig:kdr}}
    {\includegraphics[width=0.48\textwidth]{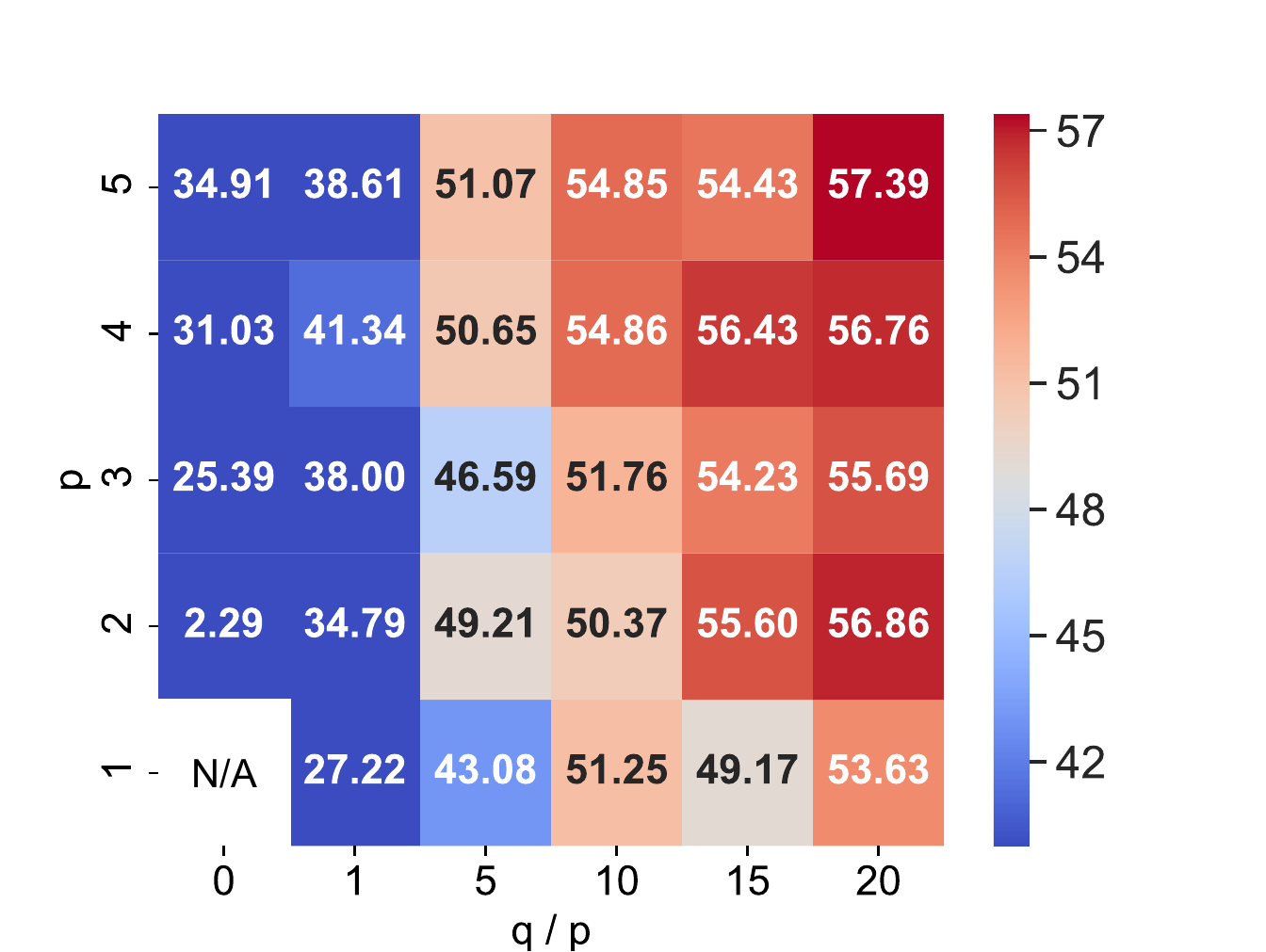}\label{fig:kdkl}}
    \vspace{-0.1in}
    \caption{Link prediction performance measured by Hits@20 of \kdr (left) and \kdkl (right) on \pubmed with different $p$ and $q$.}
    \label{fig:ablation}
\end{figure*}

\begin{table}[t]
\centering
\caption{Link prediction performance measured by Hits@20 of \ours with different fixed-length and repeating times of random walks to sample the nearby nodes for the context nodes on \cs. The default setting is 3 of 3-step random walks for the nearby nodes ($p=9$) and 120 randomly sampled nodes ($q/p=20$). RW indicates the random walk.}
\label{tab:sensitiveness_sampling}
\vspace{-0.1in}
\scalebox{0.9}{
\begin{tabular}{l|ccccc}
\toprule
\rowcolor{gray!20} RW Length                & 1 & 3 & 5  & 10 & 20 \\
Hits@20                   & 64.7      & 67.82      & 66.18     & 66.67      & 67.1       \\\hline
\rowcolor{gray!20} Repeat RW Times           & 1 & 2 & 3  & 4  & 5  \\
Hits@20                   & 65.89   &66.23     & 67.82     & 67.43      & 67.52     \\
\bottomrule
\end{tabular}}
\end{table}

\subsection{Analysis of Context Nodes Selection Strategy.}
\label{sec:apdx-context_node_selection}
\textbf{Sensitivity Analysis of $p$ and $q$ for \kdr and \kdkl.} To analyze the influence of the context nodes on \kdr and \kdkl, we plot two heat maps to show their individual performance on Pubmed under the transductive setting, as shown in \cref{fig:ablation}. These two figures show different patterns with the context nodes. \kdkl (left figure) shows that the performance becomes better with more nearby nodes ($p$) and a higher random sampled rate($q$/$p$). And random sampling rate can lead to a much better performance than nearby nodes. However, in \kdr, we find that the results on the diagonal perform consistently better than those around, which means the random sampling rate should match the nearby nodes to work well. Besides, we can also observe that \kdr is more sensitive with a smaller number of context nodes than \kdkl. \kdr matches the performance by the relative ranking of the context nodes w.r.t. the anchor nodes. However, it becomes difficult for \kdr to learn well when there are many context nodes. In contrast, \kdkl matches the distribution, and more context nodes provide a clearer picture about the link-related structure around the anchor node. 

\textbf{Sensitivity Analysis on the Fixed-Length and Repeating Times of Random Walks for the Context Nodes.}
We further conduct sensitivity analysis on the fixed lengths and repeating times of random walks for the context nodes on the \cs dataset, which is shown in \cref{tab:sensitiveness_sampling}. For the fixed length of the random walk, we find that even a relatively short length can effectively retain the local structure information. Additionally, we find that repeating the process multiple times may not lead to a significant performance boost beyond thrice, since it may already cover most of the adjacent nodes within three hops. 

\textbf{Ablation Study on Different Sampling Strategies.} Moreover, we conduct an ablation study on different efficient sampling strategies for each anchor node: 1) sampling with only repeated random walks 2) using all 3-hops Neighbors as samples 3) only randomly sampling nodes from the whole graph 4) \ours (the combination of the first and third strategies). The results are shown in \cref{tab:ablation_selection}. From the table, we observe that \ours yields the best performance by combining the local samples with global random samples. We note that \ours can also incorporate other sampling strategies, where more complex ones can potentially further improve the performances. We leave this as future work.

\begin{table}[t]
\centering
\caption{Link prediction performance evaluated by Hits@20 of \ours and different context nodes sampling strategies for each anchor node on \cs.}
\label{tab:ablation_selection}
\vspace{-0.1in}
\scalebox{0.75}{
\begin{tabular}{c|c|c|c}
\toprule
Random Walk only & 3-hop Neighbors & Random Sample only & LLP \\\midrule
60.35                     & 62.23                    & 64.92                       & 67.82 \\  
\bottomrule
\end{tabular}}
\end{table}

\subsection{Analysis of \kdr, \kdkl and \lsup for \ours}
\begin{figure}[t]
    \centering
    \includegraphics[ width=0.48\textwidth]{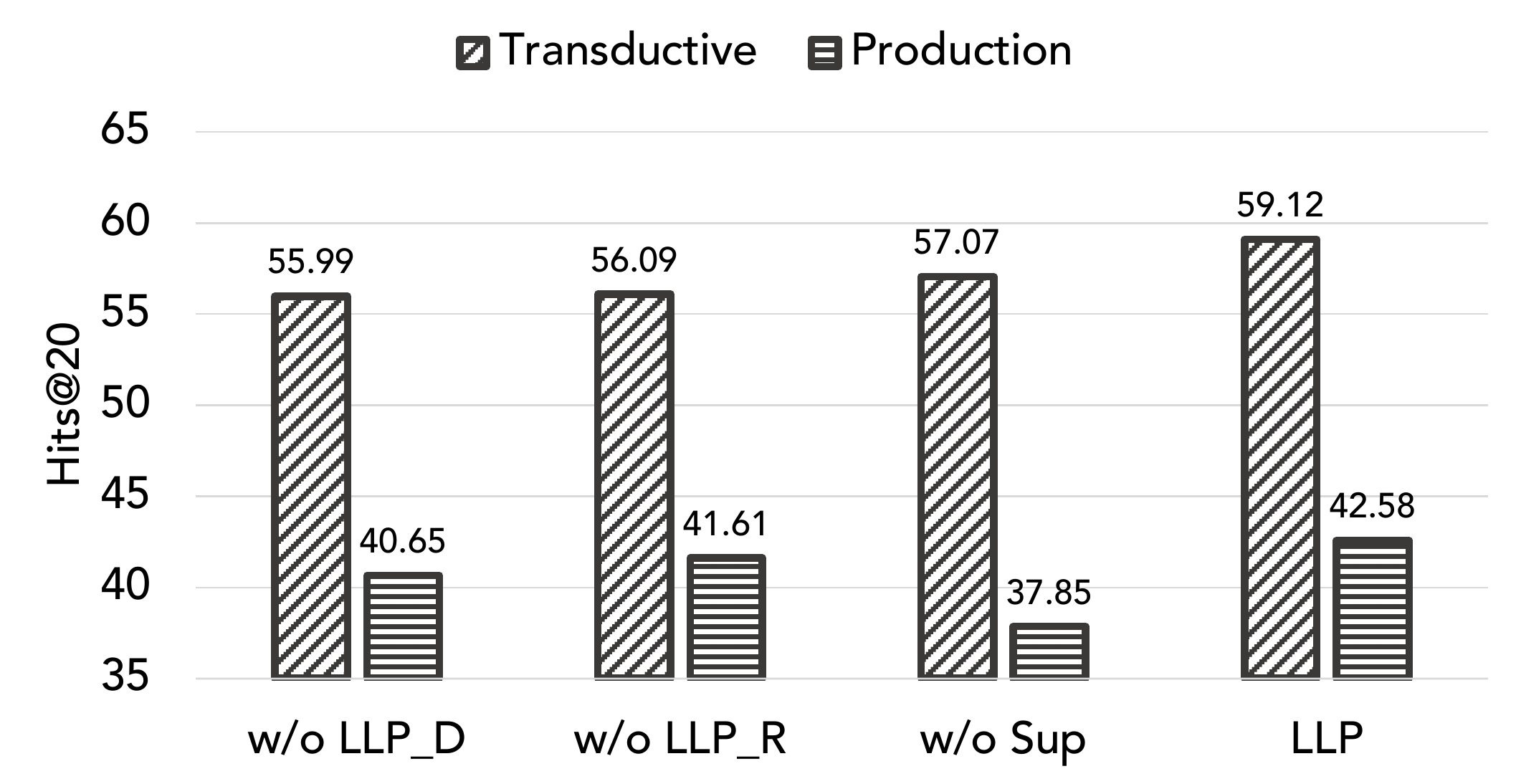}
    \caption{Link prediction performance of \ours by dropping each component in~\cref{eq:loss}. The result shown by each bar is the averaged Hits@20 across all the datasets.}
    \label{fig:ablation_overall}
\end{figure}

\textbf{Importance Analysis of \kdr, \kdkl and \lsup.}
We conduct the ablation study on all non-OGB datasets to analyze the contributions of each component in~\cref{eq:loss}. In each ablation setting, we remove one component independently, as shown in~\cref{fig:ablation_overall}. We can observe that the performance drops when any of the three components (i.e., \kdr, \kdkl, and \lsup) is removed, which shows the importance of each component. It also demonstrates that \kdr and \kdkl indeed provide complementary link prediction-related information for the student. Other than these two components, we find that the true link label information also contributes, especially under the production setting. In the production setting, as the neighbor information is sparse or absent, the limited true label information becomes critically important.

\textbf{Sensitivity Analysis of $\alpha$, $\beta$, $\gamma$ in \ours.} 
\cref{tab:sensitiveness_4params} presents the sensitiveness experiments about $\alpha$, $\beta$, and $\gamma$ on \cs dataset. We can observe that \ours is generally robust to the choices of these parameters. We also note that \ours is slightly more sensitive to $\alpha$ and $\gamma$, which means the results more rely on the ground-truth label loss and distribution-based matching loss training on \cs dataset.

\begin{table}[t]
\centering
\caption{Link prediction performance measured by Hits@20 of \ours with different $\delta$, $\alpha$, $\beta$, and $\gamma$ on \cs.}
\label{tab:sensitiveness_4params}
\scalebox{0.81}{
\begin{tabular}{l|ccccccc}
\toprule
\rowcolor{gray!20} $\delta$ & 0 & 0.001 & 0.05 & 0.1 & 0.15 & 0.2 & 0.5 \\ 
Hits@20      & 64.11      & 65.97          & 67.06         & 68.62        & 67.94         & 67.41        & 67.01        \\ \hline
\rowcolor{gray!20} $\alpha$         & 0.001      & 0.01           & 0.1           & 1            & 10            & 100          & 1000         \\ 
Hits@20      & 63.8       & 65.1           & 64.17         & 65.43        & 68.62         & 65.59        & 58.48        \\ \hline
\rowcolor{gray!20} $\beta$          & 0.001      & 0.01           & 0.1           & 1            & 10            & 100          & 1000         \\ 
Hits@20      & 67.18      & 67.85          & 68.62         & 67.96        & 67.48         & 67.77        & 65.9         \\ \hline
\rowcolor{gray!20} $\gamma$         & 0.001      & 0.01           & 0.1           & 1            & 10            & 100          & 1000         \\ 
Hits@20      & 57.45      & 55.47          & 57.41         & 57.86        & 62.73         & 68.62        & 63.43        \\ 
\bottomrule
\end{tabular}}
\end{table}

\subsection{Analysis of $\delta$ in \kdr}
\label{sec:apdx-delta}
\textbf{Necessity Analysis of $\delta$ in \kdr.} To analyze the necessity of $\delta$ in \kdr, we conduct experiments on all non-OGB datasets to compare the results using \kdr with and without $\delta$. The results are shown in~\cref{tab:delta}. We observe that the results of \kdr without $\delta$ always approach zero after several training epochs. It demonstrates the effectiveness of $\delta$ in avoiding noise and transferring useful knowledge to the student.
\begin{table*}[t]
\centering
\caption{Link prediction performance measured by Hits@20 of \kdr with and without $\delta$.}
\label{tab:delta}
\vspace{-0.1in}
\scalebox{0.85}{
\begin{tabular}{l|c|c|c|c|c|c|c} 
\toprule
Method &\cora &\citeseer &\pubmed &\cs &\physics &\computers &\photos \\
\midrule
\kdr &76.52 &75.23 & 53.30 & 68.30 &60.28 &25.98 &33.33 \\
\kdr w/o $\delta$ &0.00 &0.00 &0.00 &0.00 &0.00 &0.00 &0.00  \\
\bottomrule
\end{tabular}}
\end{table*}


\textbf{Sensitivity Analysis of $\delta$ in \ours.}
We conduct the sensitivity experiments about $\delta$ on \cs dataset. The results are shown in \cref{tab:sensitiveness_4params}. We can observe that \ours is generally robust to the choices of this parameter. 

\subsection{Further Comparison Results on Cold Start Nodes}
\label{sec:cold-start}
We further compare \ours with another related work~\citep{alon2020bottleneck}, which does not only rely on the connection relationship in the graph either. This paper identifies a bottleneck of graph neural networks, and proposes to modify the last layer to be a fully-adjacent layer (FA) as a simple strategy to circumvent the bottleneck problem. To evaluate its performance in cold-start setting, we modify the last layer of GNNs with a fully-adjacent layer following it. The results compared with \ours are shown in~\cref{tab:FA}. We observe that this method is not suitable for large datasets – it results in a dense $N\times N$ adjacency matrix and results in each node receiving messages from 
$N$ other nodes (which is problematic when $N$ is large). Most datasets we utilized in our paper with a fully-adjacent layer can not fit into an NVIDIA A100 GPU (40GB memory). The performance of GNN+FA on \cora and \citeseer is even worse than vanilla GNNs. The potential reason is link prediction tasks do not heavily depend on long-range information, where the best results were found using only 2-3 layers - we observe that~\citet{alon2020bottleneck} focuses evaluation on smaller datasets that are sensitive to long-range dependencies, which seems to be a mismatch with our intended setting. 
\begin{table*}[t]
\centering
\caption{Link prediction performance measured by Hits@20 of \ours and GNN+FA~\citep{alon2020bottleneck} on cold-start nodes.}
\label{tab:FA}
\vspace{-0.1in}
\scalebox{0.85}{
\begin{tabular}{l|c|c|c|c|c|c|c}
\toprule
Method & \cora & \citeseer & \pubmed & \cs & \physics & \computers & \photos \\
\midrule
GNN & 6.39 & 11.04 & 4.63 & 9.46 & 5.46 & 1.53 & 0.87\\
\midrule
GNN+FA & 2.03 & 2.89 & OOM & OOM & OOM & OOM & OOM \\
\midrule
\ours & 22.01 & 32.09 & 37.68 & 46.83 & 39.37 & 14.64 & 23.79\\
\bottomrule
\end{tabular}}
\end{table*}

\subsection{Comparison between \ours and Other Distillation/Rank-based Matching Methods}
\label{sec:apdx-distillation-baselines}

\begin{table}[t]
\centering
\caption{Link prediction performance measured by Hits@20 of \ours and distillation methods for other link prediction tasks.}
\label{tab:rankdistil}
\vspace{-0.1in}
\scalebox{0.85}{
\begin{tabular}{l|c|c|c}
\toprule
           & RankDistill & Topology Distillation & \ours   \\\midrule
\cora      & 74.29       & 75.51                 & 78.82 \\\midrule
\citeseer  & 70.44       & 71.47                 & 77.32 \\\midrule
\pubmed    & 39.28       & 37.89                 & 57.33 \\\midrule
\cs        & 44.55       & 60.16                 & 68.62 \\\midrule
\physics   & 49.11       & 55.32                 & 72.01 \\\midrule
\computers & 15.64       & 23.89                 & 35.32 \\\midrule
\photos    & 28.75       & 38.87                 & 49.32\\
\bottomrule
\end{tabular}}
\end{table}

\textbf{Compared with Other Distillation Methods.} Different from the link prediction tasks like RankDistill~\citep{reddi2021rankdistil} and Topology Distillation~\citep{kang2021topology} which distill the information in a non-graph context to keep the rank information among entities, we distill in a graph context to keep the graph structure information. \textbf{For the sampling methods,} RankDistill samples the nodes to teach the student based on the teacher’s results. Same for Topology Distillation, which assigns different entities into different groups and builds a hierarchical topology structure accordingly to distill the topology information. Different from these two methods, \ours samples node pairs independently of the teacher, which is only based on the graph structure. \textbf{For the matching methods,} RankDistill keeps the order for “top-K” items and penalizes high scores by the student for “bottom-K” items. Topology Distillation generates fully connected graphs based on entities or groups, where the adjacency matrix is built by the similarity among nodes in the generated graphs. Topology Distillation matches the adjacency matrix generated by the teacher and the student to preserve the topology knowledge. Differently, our method matches both the order and the distribution of the sampled node pairs between teacher and student, which we show are both useful and complementary~\cref{tab:ablation_rd} in preserving structure information distillation between the two.

We also conduct experiments to evaluate the impact of the different methods in our prediction task. For Topology Distillation, we pick their full topology distillation (FTD) method for the evaluation, as the authors demonstrate the student model takes more benefits from FTD when there is no significant capacity gap between the teacher and the student. The results are shown in \cref{tab:rankdistil}. We can observe that \ours consistently outperforms RankDistill and Topology Distillation. We believe this is because our sampling and matching methods based on the graph topology structure are more effective in preserving relevant graph structure and graph link prediction-related knowledge than the distillation methods proposed for item recommendation.

\textbf{Compared with Other Rank-based Matching Methods.} While there exist multiple techniques that can be used to implement the rank-based matching for our task, we note that our task (rank-based matching for preserving graph structural information) is not the same as other rank-matching tasks (i.e. KD for recommendation tasks). Therefore, existing rank-matching methods might not be suitable for KD in link prediction. In \cref{tab:listnet}, we compare ListNet~\citep{cao2007learning} with our proposed rank-based matching \kdr. We can observe that \kdr always outperforms ListNet across all datasets. This observation is consistent with the findings presented in \cref{sec:apdx-delta}: the conventional margin-based ranking loss cannot work well on our tasks. This is also the reason we introduce the term 
$\delta$ into our rank-matching loss~\cref{eq:loss_r}. The existing rank-based matching methods match the strict ranking order for all candidate items, as needed by the task of recommendation. But for our use case of preserving graph structural information, our proposed \kdr is more suitable due to its ability to not differentiate similar nodes within the same hop.

\begin{table}[t]
\centering
\caption{Link prediction performance measured by Hits@20 of \ours and ListNet~\citep{cao2007learning}.}
\label{tab:listnet}
\vspace{-0.1in}
\scalebox{0.85}{
\begin{tabular}{l|c|c}
\toprule
           & ListNet & \kdr \\\midrule
\cora      & 72.64   & 76.52  \\\midrule
\citeseer  & 59.82   & 75.23  \\\midrule
\pubmed    & 18.79   & 53.3   \\\midrule
\cs        & 43.06   & 68.3   \\\midrule
\physics   & 28.87   & 60.28  \\\midrule
\computers & 17.16   & 25.98  \\\midrule
\photos    & 25.68   & 33.33 \\
\bottomrule
\end{tabular}}
\end{table}

\vspace{-0.15in}
\section{Implementation Details}
\label{apdx:hyperparameter}
\textbf{Transductive Setting.}
Inspired by GLNN~\citep{zhang2021graph}, we enlarge the size of the student MLP in our experiment. As suggested by GLNN, this can significantly shorten the gap between the student MLP and the teacher GNN without greatly reducing the timing performance. We set the hidden dimension of student MLP two times larger than the teacher for \physics, \computers, and \photos, and set it four times larger than the teacher for \collab and \citationtwo. We examine the timing performance of the enlarged students by repeating the inference task ten times. The inference time of \ours increases from 1.9 to 7.1 ms on \collab and from 2.9 to 15.2 ms on \citationtwo, but it is still 18.9$\times$ and 147$\times$ faster than SAGE, respectively.

\textbf{Hyper-parameters.} We take a 2-layer SAGE (hidden size is set to 256) as the teacher for most datasets. For \citationtwo, \igbtiny, and \igbsmall, we set the layer size as 3 for better prediction results. We also take 3-layer MLP as the student on these datasets. For \ours, we conduct the hyperparameter search of the weights for \lsup, \kdr and \kdkl from [0.001, 0.01, 0.1, 1, 10, 100, 1000], the number of the nearby nodes $p$ from [1,2,3,4,5], the random sampling rate $q / p$ from [1, 3, 5, 10, 15], the learning rate from [0.001, 0.005] and the dropout rate from [0, 0.5]. 

\textbf{Implementation and Hardware Details.} Our implementation is based on PyTorch Geometric~\citep{fey2019fast}. We conduct experiments with NVIDIA V100 GPU(16GB memory). For \citationtwo and IGB datasets, we run the experiments on NVIDIA A100 GPU with 40GB memory.

\end{document}